\documentclass{article}

\usepackage{microtype}
\usepackage{graphicx}
\usepackage{subcaption}
\usepackage{multirow}
\usepackage{booktabs} 
\usepackage[table]{xcolor}
\usepackage{hyperref}

\usepackage[preprint]{icml2026}

\usepackage{enumitem}
\usepackage{amsmath}
\usepackage{amssymb}
\usepackage{mathtools}
\usepackage{amsthm}

\usepackage[capitalize,noabbrev]{cleveref}

\theoremstyle{plain}

\theoremstyle{definition}

\theoremstyle{remark}

\usepackage[textsize=tiny]{todonotes}

\icmltitlerunning{Language as a Latent Variable for Reasoning Optimization}

\begin{document}

\twocolumn[
  \icmltitle{Language as a Latent Variable for Reasoning Optimization}



  \icmlsetsymbol{equal}{*}

  \begin{icmlauthorlist}
    \icmlauthor{Linjuan Wu\footnotemark[1]}{yyy}
    \icmlauthor{Haoran Wei}{comp}
    \icmlauthor{Jialong Tang}{comp}
    \icmlauthor{Shuang Luo}{comp}
    \icmlauthor{Baosong Yang}{comp}
    \icmlauthor{Yongliang Shen}{yyy}
    \icmlauthor{Weiming Lu}{yyy}
  \end{icmlauthorlist}

  \icmlaffiliation{yyy}{Zhejiang University}
  \icmlaffiliation{comp}{Tongyi Lab, Alibaba Grou}

  \icmlcorrespondingauthor{Weiming Lu}{luwm@zju.edu.cn}

  \icmlkeywords{Machine Learning, ICML}

  \vskip 0.3in
]

\renewcommand{\thefootnote}{\fnsymbol{footnote}}
\footnotetext[1]{Work done during internship at Tongyi Lab.}



\printAffiliationsAndNotice{}  

\begin{abstract}
As LLMs reduce English-centric bias, a surprising trend emerges: non-English responses sometimes outperform English on reasoning tasks. We hypothesize that language functions as a latent variable that structurally modulates the model’s internal inference pathways, rather than merely serving as an output medium. To test this, we conducted a \textbf{Polyglot Thinking Experiment}, in which models were prompted to solve identical problems under language-constrained and language-unconstrained conditions. Results show that non-English responses often achieve higher accuracy, and the best performance frequently occur when language is unconstrained, suggesting that multilinguality broadens the model’s latent reasoning space. Based on this insight, we propose \textbf{polyGRPO} (Polyglot Group Relative Policy Optimization), an RL framework that treats language variation as an implicit exploration signal. It generates polyglot preference data online under language-constrained and unconstrained conditions, optimizing the policy with respect to both answer accuracy and reasoning structure. Trained on only ~18.1K multilingual math problems without chain-of-thought annotations, polyGRPO improves the base model (Qwen2.5-7B-Instruct) by +6.72\% absolute accuracy on four English reasoning testset and +6.89\% in their multilingual benchmark. Remarkably, it is the only method that surpasses the base LLM on English commonsense reasoning task (+4.9\%), despite being trained solely on math data—highlighting its strong cross-task generalization. Further analysis reveals that treating language as a latent variable expands the model’s latent reasoning space, yielding consistent and generalizable improvements in reasoning performance.
\end{abstract}

\section{Introduction}
LLMs excel at a wide range of tasks, particularly reasoning~\cite{o1,r1}. However, they often display an English bias—achieving stronger performance with English inputs or responses~\cite{MathOctopus,MGSM,XLT,HuangM0WW22}. Recent advances suggest that this bias is diminishing. LLMs trained on more diverse corpora increasingly demonstrate strong, and in some cases superior, reasoning abilities when operating in non-English languages~\cite{MLL_Survey,MLL_Survey2,Llama3,Qwen2.5,aya23,huang_1,huang_2,mutli-thinking}.

This raises a key question: \textbf{can polyglot thinking, the ability to reason across diverse languages, serve as a latent variable for reasoning optimization?} Recent work suggests that language diversity can indeed enhance  performance on complex tasks~\cite{huang_1,polythink26}. While training-free approaches like majority voting~\cite{cross-prompt} and automatic language selection with response weighting~\cite{AutoCAP} attempt to leverage this diversity, their effectiveness is often limited by task complexity and suboptimal language selection~\cite{huang_1}. On the other hand, RL-based methods, such as PPO~\cite{PPO}, DPO~\cite{DPO}, and GRPO~\cite{r1}, have shown promise in improving LLM reasoning. However, existing RL training often relies on English-centric or English-optimized preference data~\cite{Self-improving, MAPO}, restricting models from fully exploiting polyglot thinking, particularly when English is not the most effective reasoning language.

\begin{figure*}
    \centering
    \includegraphics[width=1\linewidth]{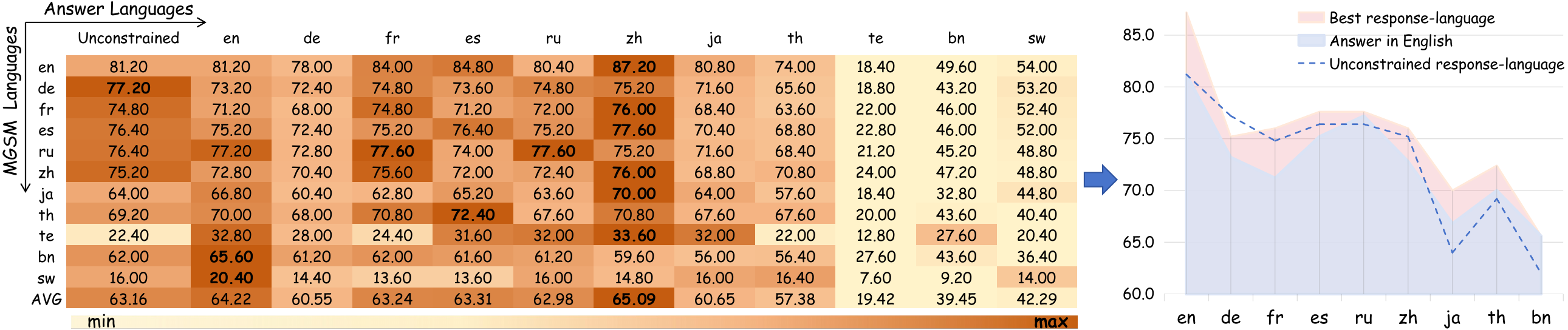}
    \caption{\textit{Polyglot Thinking Experiment} results (left part) on MGSM~\cite{MGSM} in Qwen2.5-7B-Instruct~\cite{Qwen2.5} model, including ten languages: English (en), German (de), French (fr), Spanish (es), Russian (ru), Chinese (zh), Japanese (Ja), Thai (th), Telugu (te), Bengali (Bn), and Swahili (sw). The right panel highlights the best score (red area) under specified-language settings, the score when responding in English (blue area), and the score when the response language is unconstrained (blue dashed line).}
    \label{PTE_experiments}
\end{figure*}

\begin{figure*}[t]
    \centering
    \includegraphics[width=0.8\linewidth]{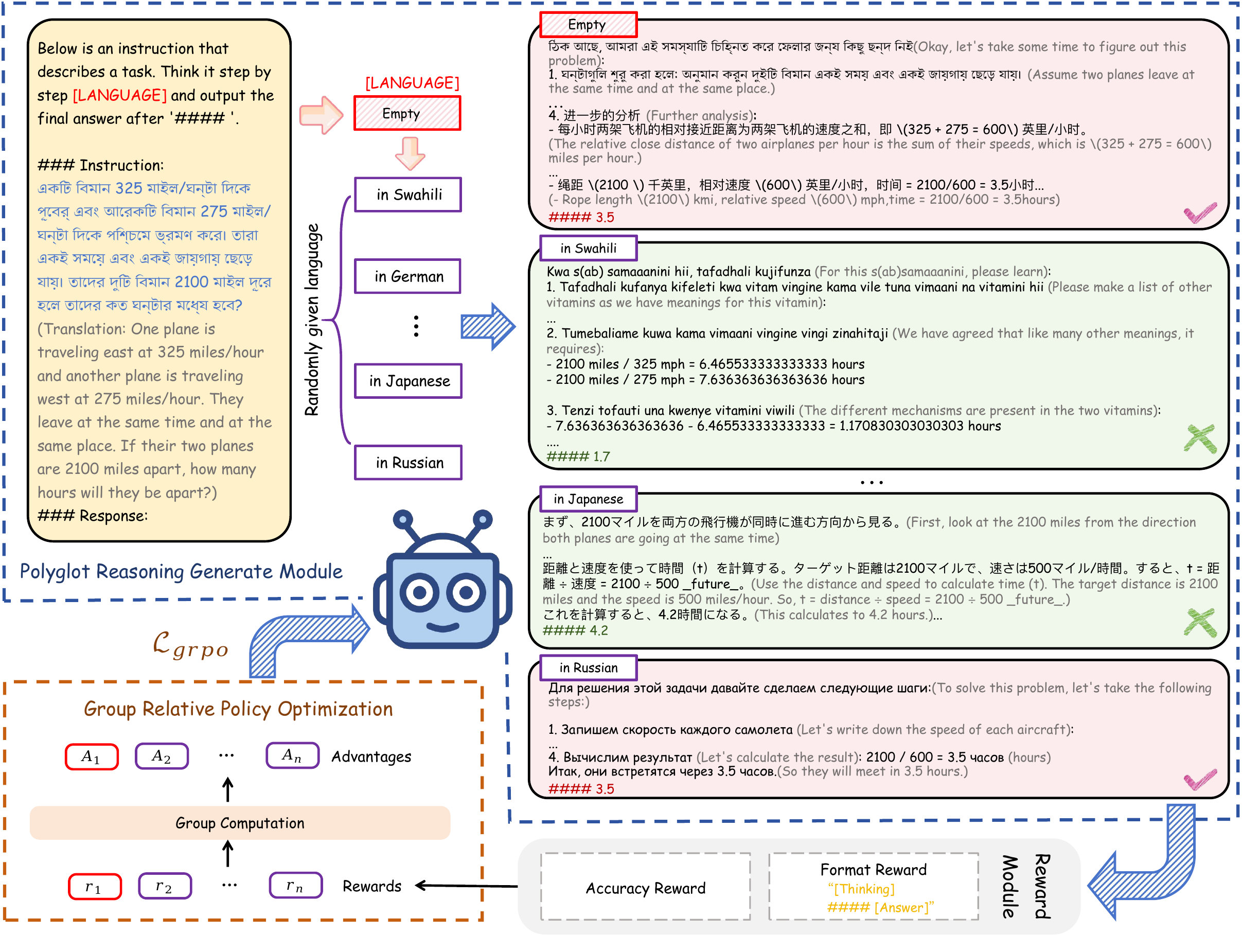}
    \caption{The framework of polyGRPO, including Polyglot Reasoning Generate Module (PRGM), Reward Module and Group Relative Policy Optimization Module.}
    \label{model}
\end{figure*}

Previous observations suggest that reasoning in non-English languages can, in certain tasks, outperform reasoning in English. To systematically investigate this effect, we introduce a \textit{Polyglot Thinking Experiment} on MGSM~\cite{MGSM}. In this setup, for each language in MGSM, we construct prompts that elicit responses under both unconstrained and constrained response-language settings. In the unconstrained setting, the model can freely choose the output language, while in the constrained setting, the output is restricted to a specified language. The detailed prompt design is illustrated in Figure~\ref{model}, and the corresponding results are shown in Figure~\ref{PTE_experiments}. Our findings indicate that, in constrained response-language settings, Chinese responses, on average, outperform English, while no single setting consistently dominates across all languages. Notably, in the unconstrained setting, models often outperform the English-only baseline. We attribute these results to a flexible reasoning space, where allowing the model to operate across multiple languages without strict language constraints expands its reasoning potential. \textbf{By treating language as a latent variable, we observe that language diversity enables the model to explore different reasoning trajectories, enhancing its ability to solve complex tasks.}

Motivated by this insight, we combine language-constrained and unconstrained prompts to form preference groups that capture polyglot reasoning trajectories. Building on GRPO~\cite{r1}, we propose polyglot GRPO (polyGRPO), a reinforcement learning method that explicitly leverages the latent reasoning space induced by polyglot thinking to enhance LLM reasoning. As shown in Figure~\ref{model}, polyGRPO consists of three components: the Polyglot Thinking Generation Module, the Reward Module, and the Group Relative Policy Optimization Module. For each question, we generate a set of responses—one under an unconstrained setting and others in randomly selected target languages—thereby creating a diverse set of reasoning paths. The reward function is rule-based, combining correctness (measured by the final answer) and format (encouraging reasoning steps). Based on these reward scores, group-relative advantages are computed to establish preference rankings, which guide policy optimization through GRPO.

We train polyGRPO using approximately 18k multilingual mathematics examples on Qwen2.5-7B-Instruct and Llama3-8B-Instruct models. Testing is conducted across four reasoning benchmarks, including MGSM~\cite{MGSM}, mMATH~\cite{MATH500}, PolyMath-medium~\cite{wang2025polymath}, and X-CSQA~\cite{XCSQA}. On average, polyGRPO improves reasoning performance over the strongest baselines by 3.07\% (on Qwen2.5), and by 1.72\% (on Llama3) in English questions. These results highlight polyGRPO’s ability to leverage polyglot reasoning for optimization.

Our contributions are summarized as follows:
\begin{itemize}[leftmargin=*]
\item We reveal that English is not always the best response language in reasoning tasks, and that unconstrained language responses often yield surprising results. Based on this insight, we propose leveraging polyglot thinking as a latent variable to improve LLM reasoning capabilities.
\item We introduce \textbf{polyGRPO}, a novel reinforcement learning framework that online generates multilingual preference sets constructed from polyglot thinking to optimize LLMs.
\item Through experiments on four reasoning benchmarks and two base models, polyGRPO significantly improves LLM performance on both mathematical and commonsense reasoning tasks, with notable gains in English reasoning, illustrating the power of polyglot thinking in enhancing reasoning capabilities without requiring polyglot inference at test time.
\end{itemize}

\section{Related Work}
\noindent \paragraph{Polyglot Thinking of LLMs.}
Early LLMs were predominantly trained on English-centric data, resulting in better performance when questions or responses were in English~\cite{MGSM}. To improve reasoning capabilities in other languages, recent work has proposed cross-lingual chain-of-thought (CoT) prompting strategies~\cite{cross-TOT,XLT}. From a training perspective, beyond merely increasing multilingual training data, some studies translate English questions~\cite{MindMerger,QAlign} or CoT responses~\cite{MathOctopus,mCoT,xCoT} into multiple languages and fine-tune models on the augmented data. Besides translation-based methods, multilingual preference training has gained traction~\cite{MAPO,Self-improving}, often treating English reasoning as the reference to guide multilingual outputs. However, as multilingual LLMs improve, their reasoning in certain languages can surpass English~\cite{huang_1}, sparking growing interest in leveraging Polyglot Thinking to enhance reasoning capabilities.

\paragraph{Enhancing LLM Reasoning with Polyglot Thinking.}
\citet{huang_1} showed that aggregating reasoning across $k$ languages (Acc@$k$) can outperform English-only reasoning by up to 10 points, with robustness to both translation quality and language selection. Building on similar insights, \citet{cross-prompt} proposed cross-lingual prompting, which first guides the model to understand the question in English before generating answers in multiple languages, with final predictions determined by majority voting. However, their method suffers from instability due to arbitrary language choices. To address this, AutoCAP~\cite{AutoCAP} introduces an automated scheme in which the LLM selects languages and assigns weights to CoTs, generating final answers through weighted multilingual outputs. Despite these efforts, such approaches remain limited by task complexity and generalization challenges.

In contrast, our method, polyGRPO, adopts a reinforcement learning framework that allows the model to internally explore and integrate Polyglot Thinking without relying on post-hoc voting or language-specific heuristics.

\section{Method}
Motivated by the diverse performance exhibited in Polyglot Thinking, we aim to let the model learn from such diversity instead of aligning all reasoning to English. We propose a polyglot reinforcement learning framework, polyGRPO, to enhance LLMs' reasoning abilities through Polyglot Thinking. As illustrated in Figure~\ref{model}, polyGRPO consists of three modules: (1) Polyglot Reasoning Generation Module (§ \ref{PRGM}), (2) Reward Module (§~\ref{RM}), and (3) Group Relative Policy Optimization Module (§~\ref{grpo}).

\subsection{Polyglot Reasoning Generation Module}\label{PRGM}

GRPO~\cite{r1} is a reinforcement learning method that improves upon PPO by removing the value function and estimating advantages in a group-relative manner. To construct the training group for a question-answer pair $(q,a) \in \mathcal{D}$, it samples $n$ responses $\{o_i\}_{i=1}^n$ from the old policy $\pi_{\theta_{\text{ref}}}$.

Our proposed \textbf{Polyglot Reasoning Generation Module (PRGM)} is designed to guide the LLM in generating a group of $n$ multilingual responses for each input, thereby exploring different reasoning trajectories across languages. As shown in the upper part of Figure~\ref{model}, given an input question, we generate a set of $n$ responses using prompts $\{p_i\}_{i=1}^n$ with or without explicit language instructions. Specifically, one response is generated with no language constraint (i.e., "$[$LANGUAGE$]$" is empty), while the remaining responses are generated using prompts that specify a reasoning language randomly chosen from a predefined set of languages. These responses form the \textit{Polyglot Thinking} set $\{o_i\}_{i=1}^n$, which may include both correct and incorrect answers.

This module operates \textbf{online during training}, enabling the model to continuously generate fresh multilingual responses. Such an online approach reduces storage overhead compared to previous methods~\cite{MAPO, Self-improving}, while facilitating broader exploration of the Polyglot Thinking space. By allowing the model to generate responses across multiple languages, PRGM expands the latent reasoning space, empowering the model to explore a more diverse set of reasoning paths, ultimately enhancing its overall reasoning capabilities.

\subsection{Reward Module}\label{RM}

Each response $o_i$ is then evaluated using a reward module to obtain an individual reward $r_i$. To assess response quality, we design a rule-based reward function composed of two parts $r_i = \text{AR}(o_i) + \text{FR}(o_i)$, where:
\begin{itemize}[leftmargin=*]
    \item \textbf{Accuracy Reward (AR):} A binary reward that evaluates whether the final predicted answer exactly matches the gold-standard answer. Formally,
    \begin{equation}
    \text{AR}(o_i) = 
    \begin{cases}
        1, & \text{if}\  \text{Answer}(o_i) = \text{Gold}(q) \\
        0, & \text{otherwise}
    \end{cases}
    \end{equation}

    \item \textbf{Format Reward (FR):} A binary reward that encourages structured reasoning. It returns 1 if the response contains a reasoning process (denoted by the keyword \texttt{[Thinking]}) and presents the final answer in the required format (i.e., following \texttt{"\#\#\#\# "}). Formally,
    \begin{equation}
    \text{FR}(o_i) = 
    \begin{cases}
        1, & \text{if} o_i \text{ contains} \texttt{[Thinking]} \\ &\text{ and \texttt{[Answer]} follows} \texttt{"\#\#\#\# "} \\
        0, & \text{otherwise}
    \end{cases}
    \end{equation}
\end{itemize}

To prevent the model from taking shortcuts, e.g., generating minimal text before directly outputting the answer, we additionally enforce a minimum length constraint of 100 characters for the reasoning content within the \texttt{[Thinking]} section as part of the format reward.

The final reward $r_i \in \{0, 1, 2\}$ thus encourages both correctness and structured reasoning format, without requiring annotated reasoning steps.

\subsection{Group Relative Policy Optimization Module}\label{grpo}
Then, we follows the GRPO~\cite{r1} to optimization our model as shown in left-downer corner of Figure~\ref{model}. The advantage $A_i$ of the $i$-th response is calculated by normalizing the rewards $\{r_i\}_{i=1}^n$ of the group:
\begin{equation}
A_i = \frac{r_i - \text{mean}(\{r_i\}_{i=1}^n)}{\text{std}(\{r_i\}_{i=1}^n)}.
\end{equation}

GRPO adopts a PPO-style clipped objective, with a KL penalty between the current policy $\pi_\theta$ and the reference model $\pi_{\theta_{\text{ref}}}$ directly integrated into the loss to simplify training.

So, the loss of our polyGRPO is:
\begin{equation}
\begin{aligned}
\mathcal{L}_{\text{polyGRPO}}& (\theta) = \,
\mathbb{E}_{(q,a) \sim \mathcal{D}, \{o_i\}_{i=1}^n \sim {\pi_{\theta_{\text{ref}}}(o_i|p_i,q)}_{i=1}^n} \Bigg [
\frac{1}{n}\sum_{i=1}^n \\
& 
\frac{1}{|o_i|}\sum_{t=1}^{|o_i|}
\Bigg\{
\min\Bigg[\frac{\pi_{\theta}^{i,t}}{\pi_{\theta_{\text{ref}}}^{i,t}}\hat{A}_i, \text{clip}\Bigg(\frac{\pi_{\theta}^{i,t}}{\pi_{\theta_{\text{ref}}}^{i,t}}, \\ & 1 - \epsilon, 1 + \epsilon\Bigg)\hat{A}_i\Bigg]
- \beta \mathbb{D}_{\text{KL}}(\pi_\theta \| \pi_{\text{ref}})
\Bigg\}\Bigg ]
\end{aligned}
\end{equation}
where $\pi^{i,t}$ denotes the conditional probability of the token at position $t$, formally:
\begin{equation}
    \pi^{i,t} = \pi(o_{i,t} |p_i,q,o_{i,<t}),
\end{equation}
where $p_i$ is the $i$-th prompt with or without explicit language instructions to obtain $o_i$.

Compared to previous approaches that rely on supervised translations~\cite{MAPO} or fixed language anchors~\cite{Self-improving}, polyGRPO allows LLMs to autonomously explore and integrate Polyglot Thinking as a latent variable, enabling a more flexible and effective reasoning process.

\begin{table*}[t]
\centering
\caption{Main experimental results for Qwen2.5-7B-Instruct and Llama3-8B-Instruct models across various benchmarks. "EN" refers to the English-only performance, while "ALL" indicates performance across multilingual evaluation.}
\resizebox{0.85\textwidth}{!}{\begin{tabular}{lcc|cc|cc|cc|cc}
    \toprule
\multirow{1}{*}{\textbf{Model}} & \multicolumn{2}{c|}{\textbf{MGSM}} & \multicolumn{2}{c|}{\textbf{MATH500}} & \multicolumn{2}{c|}{\textbf{PolyMath-medium}} & \multicolumn{2}{c|}{\textbf{X-CSQA}} & \multicolumn{2}{c}{\textbf{AVG.}} \\
\cline{2-11}
 & \textbf{EN} & \textbf{ALL.} & \textbf{EN} & \textbf{ALL.} & \textbf{EN} & \textbf{ALL.} & \textbf{EN} & \textbf{ALL.} & \textbf{EN} & \textbf{ALL.} \\
\midrule
\rowcolor[rgb]{ .906,  .902,  .902} \textit{\textbf{Qwen2.5-7B-Instruct}} & & & & & & & & & &  \\
\midrule
Base~\citep{Qwen2.5} & 81.20 & 63.16 & 70.80 & 57.89 & 26.40 & 23.69 & 77.10 & 54.27 & 63.88 & 49.75 \\
xRFT~\citep{MathOctopus,MAPO} & \textbf{94.00} & 68.47 & 73.40 & 53.12 & 28.80 & 18.40 & 70.30 & 49.25 & 66.62 & 47.31 \\
LIDR~\citep{Self-improving} & 90.00 & 69.60 & 73.20 & 62.46 & 28.00 & 25.07 & 75.10 & 53.18 & 66.57 & 52.58 \\
MAPO~\citep{MAPO} & 84.80 & 66.29 & 76.20 & 61.20 & \textbf{32.00} & 23.24 & 77.10 & 50.69 & 67.52 & 50.36 \\
GRPO~\citep{r1} & 90.40 & 73.02 & 74.80 & 63.20 & 26.40 & 23.42 & 75.50 & 57.10 & 66.78 & 54.19 \\
polyGRPO (Ours) & \textbf{94.00} & \textbf{75.93} & \textbf{76.80} & \textbf{64.94} & 29.60 & \textbf{25.24} & \textbf{82.00} & \textbf{60.46} & \textbf{70.60} & \textbf{56.64} \\
\midrule
\rowcolor[rgb]{ .906,  .902,  .902} \textit{\textbf{Llama3-8B-Instruct}} & & & & & & & & & &   \\
\midrule
Base~\citep{Llama3} & 79.60 & 52.22 & 29.20 & 26.00 & 8.00 & 4.84 & 66.30 & 45.12 & 45.78 & 32.05 \\
xRFT~\citep{MathOctopus,MAPO} & 73.20 & 53.89 & 27.40 & 23.37 & 6.40 & 4.93 & 67.70 & 48.15 & 43.67 & 32.59 \\
LIDR~\citep{Self-improving} & 79.60 & 55.53 & 28.00 & 24.31 & 7.20 & 5.02 & 69.50 & 52.22 & 46.07 & 34.27 \\
MAPO~\citep{MAPO} & 80.40 & 60.69 & 30.40 & 25.26 & 6.40 & 4.31 & 62.30 & 43.89 & 44.87 & 33.54 \\
GRPO~\citep{r1} & 80.80 & 64.58 & 30.00 & 24.43 & 8.00 & 4.76 & 68.80 & 53.36 & 46.90 & 36.78 \\
polyGRPO (Ours) & \textbf{82.00} & \textbf{68.11} &\textbf{32.00} & \textbf{26.71} & \textbf{9.60} & \textbf{5.51} & \textbf{70.90} & \textbf{53.64} & \textbf{48.62} & \textbf{38.49} \\
\bottomrule
\end{tabular}}
\label{main_results}
\end{table*}

\section{Experiments}
\subsection{Datasets} \label{data}
\paragraph{Training Datasets.}
We use the mathematical reasoning dataset from MAPO~\cite{MAPO} as training data. It consists of 1,703 English questions from a subset of NumGLUE~\cite{NumGLUE}, along with ChatGPT-translated versions in nine languages, including Bengali (BN), Thai (TH), Swahili (SW), Japanese (JA), Chinese (ZH), Russian (RU), German (DE), Spanish (ES), and French (FR), resulting in a total of 18,140 examples. The inclusion of multilingual data ensures alignment with existing approaches, such as MAPO~\cite{MAPO} and Self-improving~\cite{Self-improving}, allowing for fair comparisons and facilitating the exploration of Polyglot Thinking in the context of multilingual reasoning tasks. 

\paragraph{Benchmarks.} Our evaluation is based on three mathematical reasoning benchmarks (MGSM, MATH500, and PolyMath) and one commonsense reasoning benchmark (X-CSQA) to assess improvements in LLM reasoning capabilities. \textbf{MGSM}~\cite{MGSM} serves as an in-domain benchmark, derived from 250 GSM8K~\cite{MathOctopus} test samples translated by native speakers into 10 typologically diverse languages. \textbf{MATH500}~\cite{MATH500} is an out-of-domain benchmark consisting of 500 diverse mathematical problems in English, with six additional translated versions for multilingual evaluation. {PolyMath}~\cite{wang2025polymath} offers a multilingual reasoning benchmark with 9,000 math problems across 18 languages at four difficulty levels. We use the \textbf{PolyMath-medium} subset as a more challenging with125 problems. \textbf{X-CSQA}~\cite{XCSQA} extends CSQA to 16 languages and challenges models to interpret complex logical relations expressed across diverse linguistic forms. The total number of evaluation languages is 23, with detailed language coverage provided in Appendix~\ref{bench_lang}.

\subsection{Experimental Setup}\label{setup}

\paragraph{Base Models and Baselines.}
We evaluate polyGRPO on the Qwen2.5-7B-Instruct~\cite{Qwen2.5} and Llama3-8B-Instruct~\cite{Llama3} models. For baselines, we compare polyGRPO with several strong methods: (1) \textbf{xRFT}~\cite{xrft}, a rejection sampling-based method using CoT traces generated and translated from Qwen-Math-7B-Instruct~\cite{Qwen-Math}; (2) \textbf{MAPO}~\cite{MAPO}, which aligns Polyglot Thinking paths to English through translation-estimated-based preference optimization; (3) \textbf{LIDR}~\cite{Self-improving}, which employs self-improving DPO training based on performance disparities between non-English languages and English; and (4) \textbf{GRPO}~\cite{r1}, which uses multilingual training data but generates only English responses to isolate the impact of Polyglot Thinking on reasoning. Full training configurations and data construction details are provided in Appendix~\ref{apx:baselines}.

\paragraph{Training Details.}
For PRGM, we use a 10-language set to guide the roll-out, aligned with the languages in the training data. The roll-out is set to $n=5$, including one non-language-constrained response and four responses in randomly selected languages from the set. Training is conducted using the verl\footnote{https://github.com/volcengine/verl} RL framework. For the Qwen2.5-7B-Instruct model, polyGRPO is trained for 5 epochs with a learning rate of $1 \times 10^{-6}$ and a batch size of 256. For the Llama3-8B-Instruct model, polyGRPO is trained for 1 epoch with the same settings. All models are trained using 8 NVIDIA A100 GPUs.

\paragraph{Inference Setup.}
At inference time, we use the same prompt format as during training (as shown in the left above part of Figure~\ref{model}), leaving the language token $[$LANGUAGE$]$ empty to allow the model to freely choose its response language. Reasoning steps are generated via greedy decoding. Final answers are extracted using rule-based parsing and evaluated using accuracy as the main metric.

\subsection{Results}
We systematically evaluated polyGRPO's performance on two mainstream baseline models. Table~\ref{main_results} presents the results across four reasoning benchmarks. Our method consistently outperforms existing approaches in both reasoning performance and generalization across different difficulty levels. Per-language results for all test sets are provided in Appendix~\ref{apx:perlang}.

\paragraph{Performance on English Reasoning Tasks.}
When evaluated on the Llama3-8B-Instruct model, polyGRPO achieved state-of-the-art performance on all tasks in English compared to baselines. For the average performance across four benchmarks, polyGRPO based on Qwen2.5-7B-Instruct and Llama3-8B-Instruct outperformed the strongest baselines by 3.07\% and 1.72\%, respectively. Notably, in the previously unseen commonsense reasoning task, X-CSQA, polyGRPO trained on Qwen2.5-7B-Instruct was the only model among all baselines to show improvement over the base LLM, with a 4.9\% increase. This demonstrates that our method enhances the model's reasoning abilities, not just its mathematical reasoning capabilities.

\paragraph{Performance on Multilingual Reasoning Tasks.}
PolyGRPO also performs exceptionally well in multilingual reasoning tasks, achieving the best results across both base LLMs. In particular, on the MGSM dataset, polyGRPO outperforms the strongest baseline by 2.91\% (based on Qwen2.5) and 3.53\% (based on Llama3) in average multilingual performance. Similarly, polyGRPO achieved a 3.36\% average improvement on the X-CSQA benchmark compared to Qwen2.5-7B-Instruct, outperforming all baselines with a score of 60.46\%. These results demonstrate polyGRPO's ability to effectively leverage polyglot reasoning, expanding the model's reasoning capabilities not only for English but also across multiple languages, including low-resource languages in the test sets.

Overall, polyGRPO demonstrated significant improvements across both LLMs with different architectures, showcasing the versatility and robustness of our approach. Compared to GRPO, polyGRPO further enhanced key metrics, effectively demonstrating the impact of Polyglot Thinking in boosting model reasoning performance.

\subsection{Ablation Study}\label{sec:ablation}

We conduct ablation studies on the Qwen2.5-7B-Instruct model and evaluate it on the MGSM benchmark. First, we examine the impact of the format reward (i.e., w/o format reward). Next, we compare three PRGM roll-out variants: (1) without the unconstrained response-language roll-out (i.e., w/o unconstrained response); (2) with only the unconstrained response-language setting for roll-out (i.e., only roll-out unconstrained response); and (3) only the English response roll-out (i.e., GRPO setting). Additionally, to assess the performance of our method on smaller models, we include two other sizes of Qwen2.5-Instruct models, with 1.5B and 3B parameters, respectively. Results are shown in Figure~\ref{ablation}, with further details in Appendix~\ref{apx:ablation}.

\begin{figure}[t]
    \centering
    \includegraphics[width=1\linewidth]{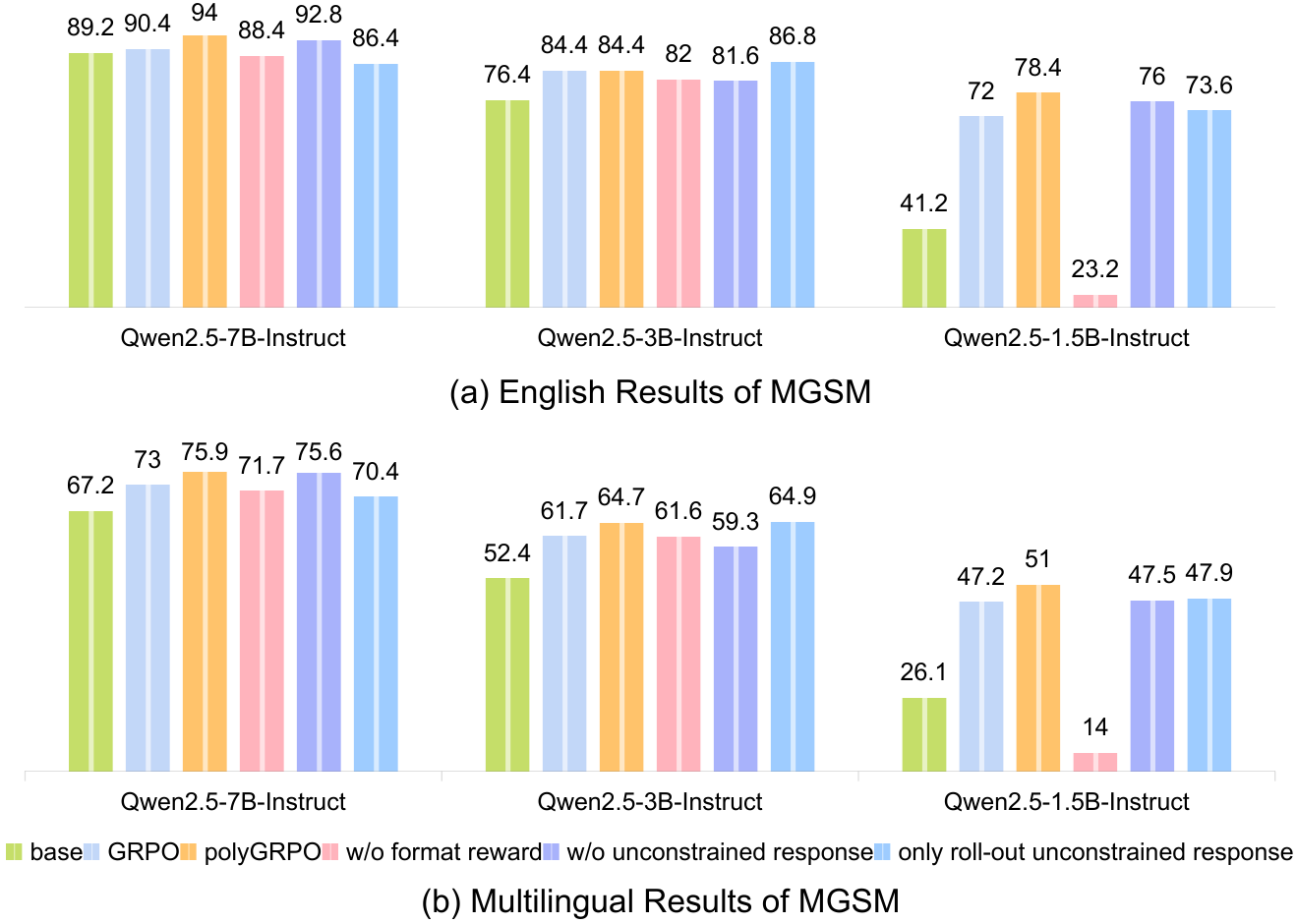}
    \caption{Ablation results on MGSM with three sizes of Qwen2.5-Instruct models.}
    \label{ablation}
\end{figure}

The ablation studies validate the importance of each setting in ensuring the effectiveness of polyGRPO. Based on the results and model behavior, we make the following observations:

\noindent \textbf{Format reward is crucial for guiding LLMs to generate Polyglot Thinking paths and valid final answers.} Without the format reward, during the roll-out phase of PRGM, the model often exhibits uncontrolled behavior, particularly in low-resource languages (e.g., Thai, Swahili), such as skipping reasoning steps, adding irrelevant text, or repeating content, specially in small size LLM (such as 1.5B model).

\noindent \textbf{Response generation with or without constrained language yields inconsistent results across LLMs of different sizes, while polyGRPO, which incorporates both settings, demonstrates more stable performance.} We observe that when all roll-outs are language-constrained, the performance of the 1.5B and 3B models drops significantly. While unconstrained language roll-outs improve performance by 2.4\% on the 3B model in English test data compared to polyGRPO, they cause a 7.6\% drop on the 7B model. Both response-language settings are essential for constructing the Polyglot Thinking roll-out.

\section{Analysis}

\subsection{The impact of language set composition.}\label{n_analysis}

\begin{figure}
    \centering
    \includegraphics[width=1\linewidth]{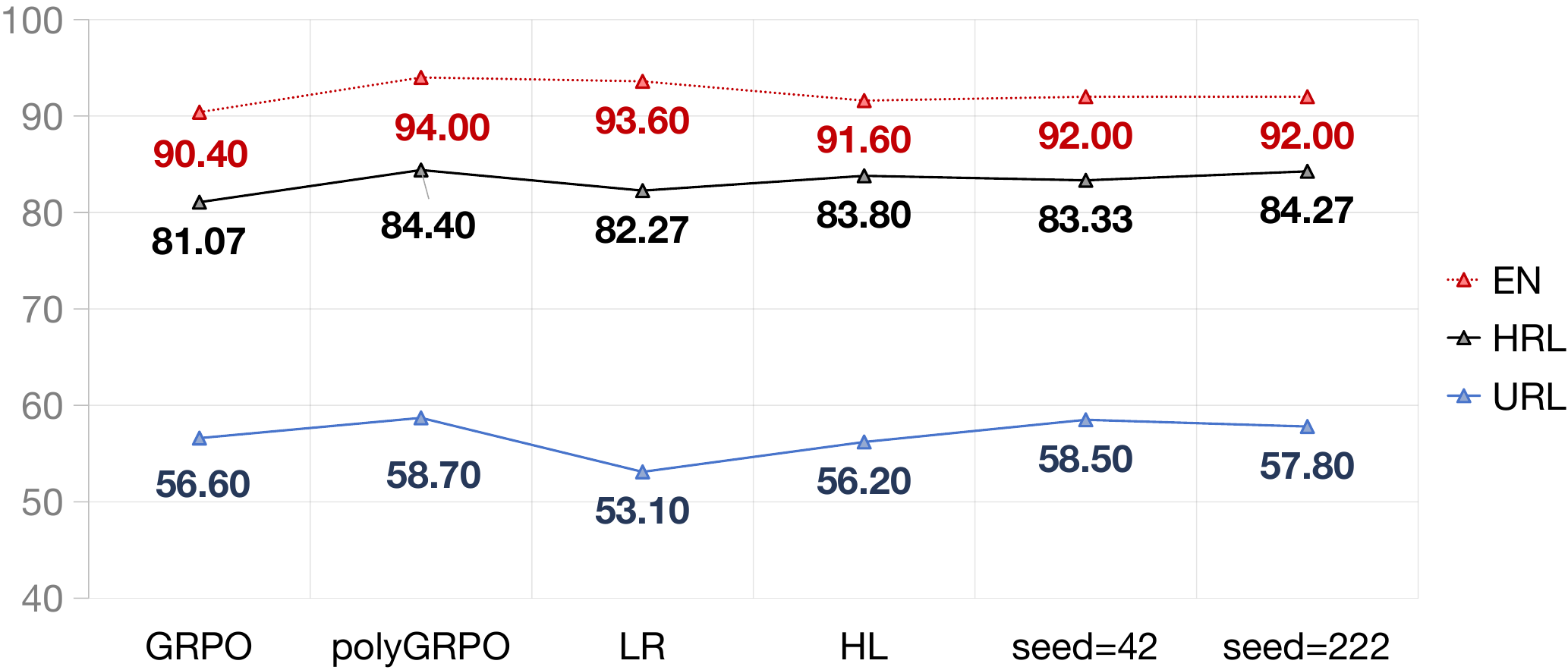}
    \caption{The results in MGSM with different language set for Polyglot Thinking base on Qwen2.5-7B-Instruct. Languages in MGSM are categorized into high-resource (HRL: ZH, FR, DE, JA, RU, ES) and underrepresented-resource (URL: BN, SW, TE (Telugu), TH) groups based on their presence in pretraining corpora such as mC4~\citep{mT5}.}
    \label{lang}
\end{figure}

The languages used in Polyglot Thinking within PRGM are primarily selected at random from a predefined language set. To align with the MAPO and LIDR methods, our language set includes all 10 languages from the training data, covering both high-resource languages (HRL) and low-resource languages (LRL, as defined in Appendix~\ref{bench_lang}). Analyzing the impact of different language sets on polyGRPO allows us to better understand the robustness and potential biases of our method with respect to language selection.

First, we established two language sets, low-resource (LR) and high-resource (HR) languages, each consisting of 10 languages and determined by their presence in pretraining corpora (such as mC4~\citep{mT5} mentioned in MGSM). The LR set includes Bengali (BN), Thai (TH), Swahili (SW), Telugu (TE), Vietnamese (VI), Basque (EU), Arabic (AR), Hindi (HI), Urdu (UR), and Turkish (TR). The HR set includes Italian (IT), Chinese (ZH), English (EN), French (FR), German (DE), Japanese (JA), Russian (RU), Spanish (ES), Korean (KO), and Portuguese (PT). Models trained with HR or LR languages are referred to as polyGRPO w/ HR (simple HR in figure) and polyGRPO w/ LR (simple LR in figure), respectively. Second, to test the robustness of polyGRPO with a mix of HR and LR languages, we created two additional sets (random seeds 42 and 222), each randomly sampling 5 HR and 5 LR languages: (1)Seed=42: [EN, KO, ZH, DE, FR, VI, BN, TR, TH, AR]; (2)Seed=222: [FR, ES, IT, KO, JA, SW, TE, VI, BN, UR].

The results on MGSM are shown in Figure~\ref{lang}. In comparison to the GRPO setup with English-only roll-out, polyGRPO consistently outperforms, regardless of the language set used. For both English  and HRL, polyglot thinking contributes positively to reasoning performance. For URL, we observe that the performance difference across various language sets diminishes when both high-resource and low-resource languages are included. This suggests that the inclusion of both language categories reduces the impact of language selection on model performance. However, the set with significant differences (seed=222) showed a slight overall performance decline, likely due to the omission of the primary language (EN) and the high-resource language (ZH), which are crucial for guiding the low-resourc language reasoning process.

We further experiment on the impact of different language quantities and roll-out values on performance, with results reported in the Appendix~\ref{apx:l_set_num} and ~\ref{apx:n}, respectively.

\subsection{Polyglot Thinking during Training}
To investigate how the model follows "Polyglot Thinking" prompts, we track the language adherence score, measured as a binary score ("0" or "1"), throughout training. In the unconstrained setting, the adherence score is set to "1" by default. We observed that as training progresses, the model gradually shifts to English-only reasoning, resembling GRPO's behavior. To further investigate this transition, we extended the training from 5 to 10 epochs (700 steps total), using the Qwen2.5-7B-Instruct model.

\subsubsection{Language Adherence of PRGM}
\begin{figure}[t]
    \centering
    \includegraphics[width=1\linewidth]{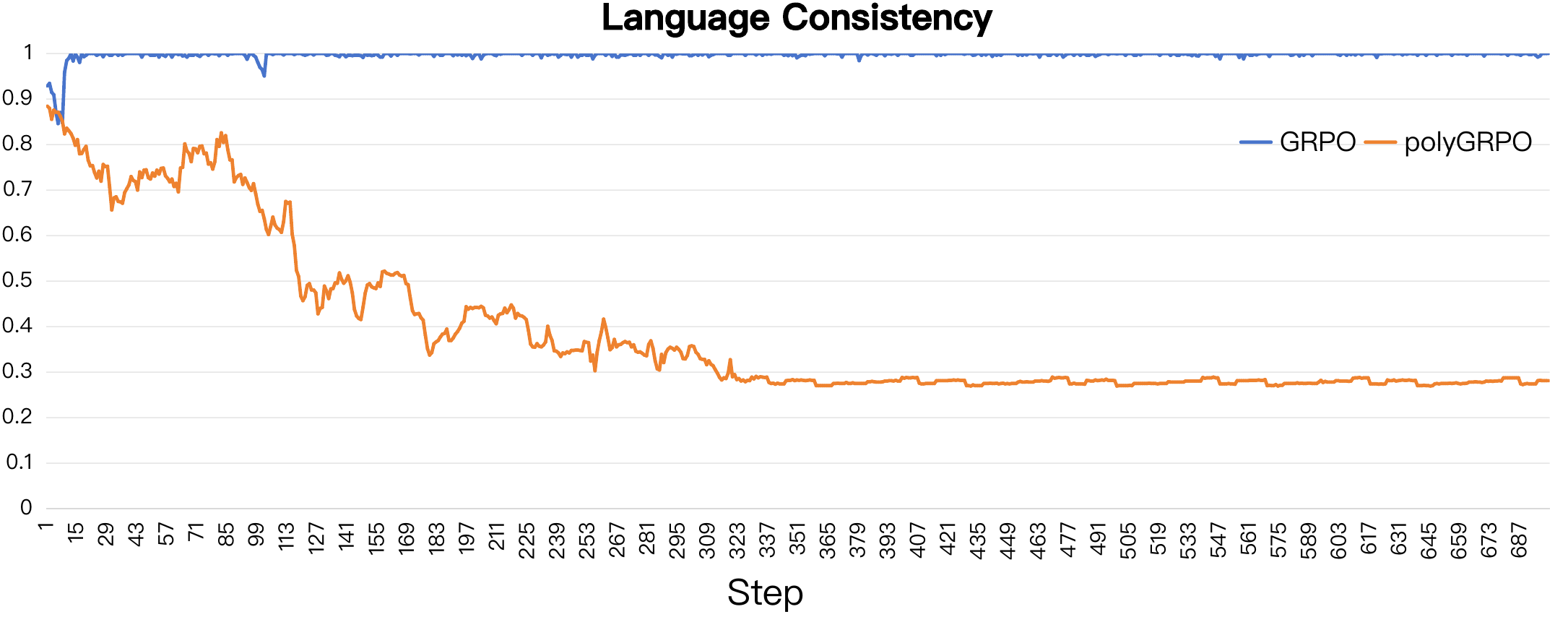}
    \caption{The language consistency of GRPO and polyGRPO during training process based on Qwen2.5-7B-Instruct model with 10 epochs (700 steps total).}
    \label{LC}
\end{figure}

As shown in Figure~\ref{LC}, GRPO quickly converges to English-only reasoning, despite some initial polyglot responses (e.g., Chinese, Swahili). In contrast, polyGRPO starts with high adherence with a gradually decreases, allowing more optimization through Polyglot Thinking. By epoch 5, the model predominantly generates reasoning in English, indicating a shift from multilingual to English-dominant reasoning.

\begin{table}[t]
\centering
\caption{The response languages for TH and BN on training data  with unconstrained-languages prompt.}
\label{tab:th_bn_performance}
\resizebox{0.45\textwidth}{!}{\begin{tabular}{l|c|c}
\hline
\multirow{2}{*}{\textbf{MGSM}} & \multicolumn{1}{c|}{\textbf{TH }} & \multicolumn{1}{c}{\textbf{BN }} \\
 & response languages & response languages \\
\hline
Qwen2.5-7B-Instruct & zh: 59.18\% th: 40.82\% & en: 90.01\% zh: 8.39\% bn: 1.6\% \\
polyGRPO \ \ \ \ \ \ epoch=1 & zh: 82.38\% en: 17.20\% & zh: 67.73\% en: 31.90\% \\
      \multicolumn{1}{r|}{epoch=2} & zh: 28.13\% en: 71.81\% & zh: 2.71\% en: 96.87\% \\
      \multicolumn{1}{r|}{epoch=3} & en: 99.70\% & en: 99.82\% \\
      \multicolumn{1}{r|}{epoch=4} & en: 99.76\% & en: 99.88\% \\
      \multicolumn{1}{r|}{epoch=5} & en: 99.76\% & en: 99.70\% \\
      \multicolumn{1}{r|}{epoch=6} & en: 99.64\% & en: 99.94\% \\
      \multicolumn{1}{r|}{epoch=7} & en: 99.76\% & en: 99.82\% \\
      \multicolumn{1}{r|}{epoch=8} & en: 99.68\% & en: 99.82\% \\
      \multicolumn{1}{r|}{epoch=9} & en: 99.94\% & en: 100.00\% \\
      \multicolumn{1}{r|}{epoch=10} & en: 99.82\% & en: 99.88\% \\
\hline
\end{tabular}}
\end{table}

\begin{figure}[t]
    \centering
    \includegraphics[width=0.82\linewidth]{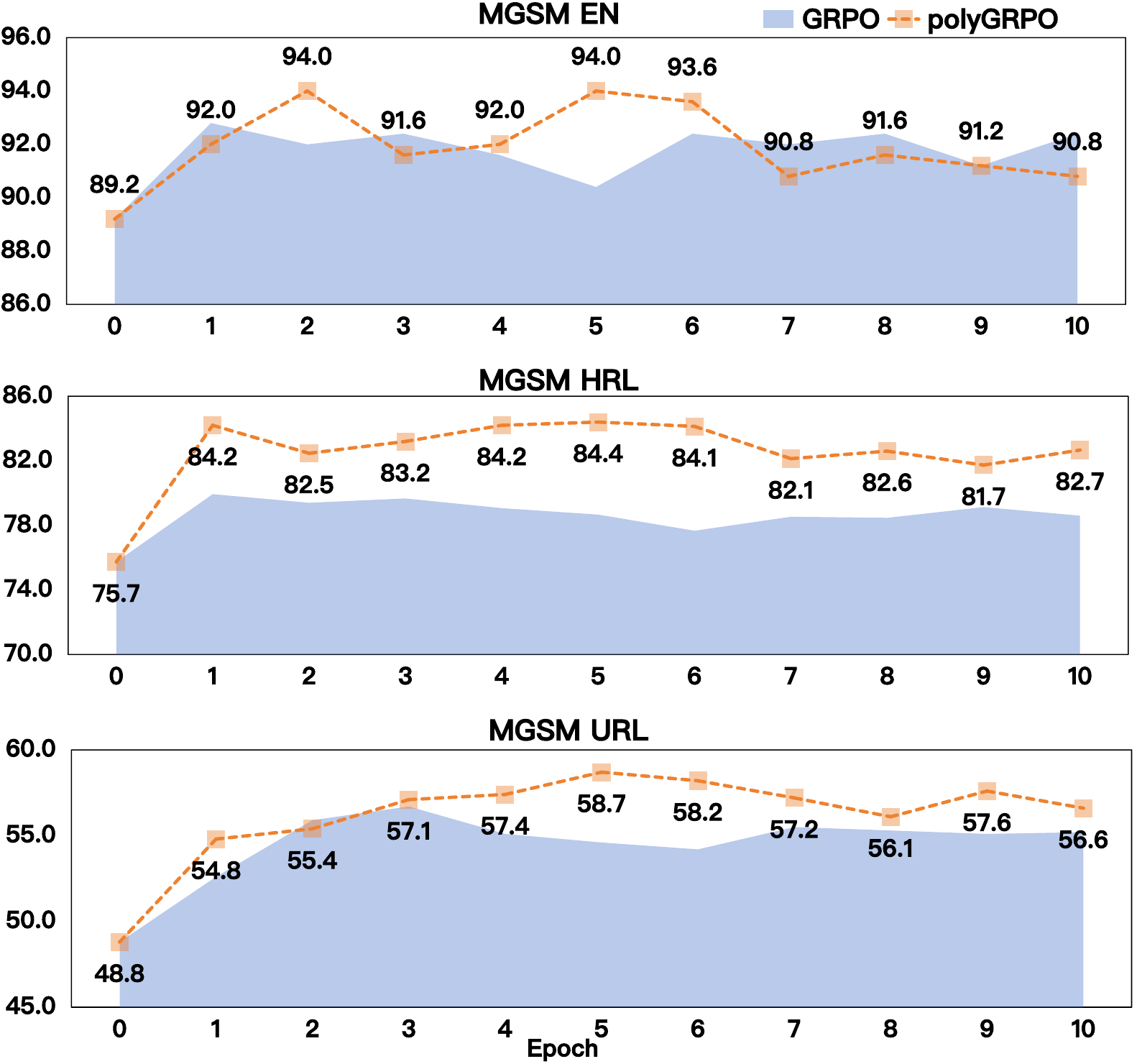}
    \caption{The performance of GRPO and polyGRPO in MGSM benchmark with 10 epochs checkpoints trained based on Qwen2.5-7B-Instruct model.}
    \label{acc}
\end{figure}

Under the unconstrained setting, the model initially generates mixed-language responses (e.g., 47.18\% EN, 42.17\% ZH in epoch 1), quickly shifting to English-dominant output by epoch 2 (82.70\%), and predominantly English thereafter. A detailed analysis of Thai (TH) and Bengali (BN) performance (Table~\ref{tab:th_bn_performance}) reveals similar dynamics: the base model demonstrates Polyglot Thinking (e.g., Chinese/Thai for Thai questions, English/Chinese for Bengali), with code-switching, while polyGRPO transitions from multilingual (base and epoch 1) to stable English-only reasoning. This process illustrates the consolidation of polyglot exploration into a unified, English-centric reasoning strategy.

Our evaluation is conducted under unconstrained-language prompts. While English predominates in the final reasoning outputs, code-switching persists, with terms or entities from the question-language integrated into the responses. This highlights that, even when primarily generating English output, \textbf{polyGRPO retains and effectively leverages a flexible Polyglot Thinking space during inference}.

\subsubsection{Performance Evolution with Extended Training}
Figure~\ref{acc} illustrates the performance of GRPO and polyGRPO on the MGSM benchmark with 10 epochs of training using the Qwen2.5-7B-Instruct model. For English (EN) performance, polyGRPO shows an initial improvement over GRPO, with a peak performance reached by epoch 5. However, after this point, the performance begins to decline, likely due to the model focusing on generating English-only responses during the later epochs, which may limit further improvement. In contrast, polyGRPO maintains consistent improvement in HRL, outperforming GRPO across the training epochs. For URL, polyGRPO shows initial gains but, like GRPO, the performance stabilizes after a few epochs. This highlights that polyglot thinking has long-term benefits in non-English questions, especially for high-resource languages. \textbf{It indicates that Polyglot Thinking in early training establishes a stronger foundation and effectively enhances the model’s reasoning capability.} This aligns perfectly with our motivation.

\subsubsection{Performance Dynamics and Entropy Analysis}

\begin{figure}[t]
    \centering
    \includegraphics[width=1\linewidth]{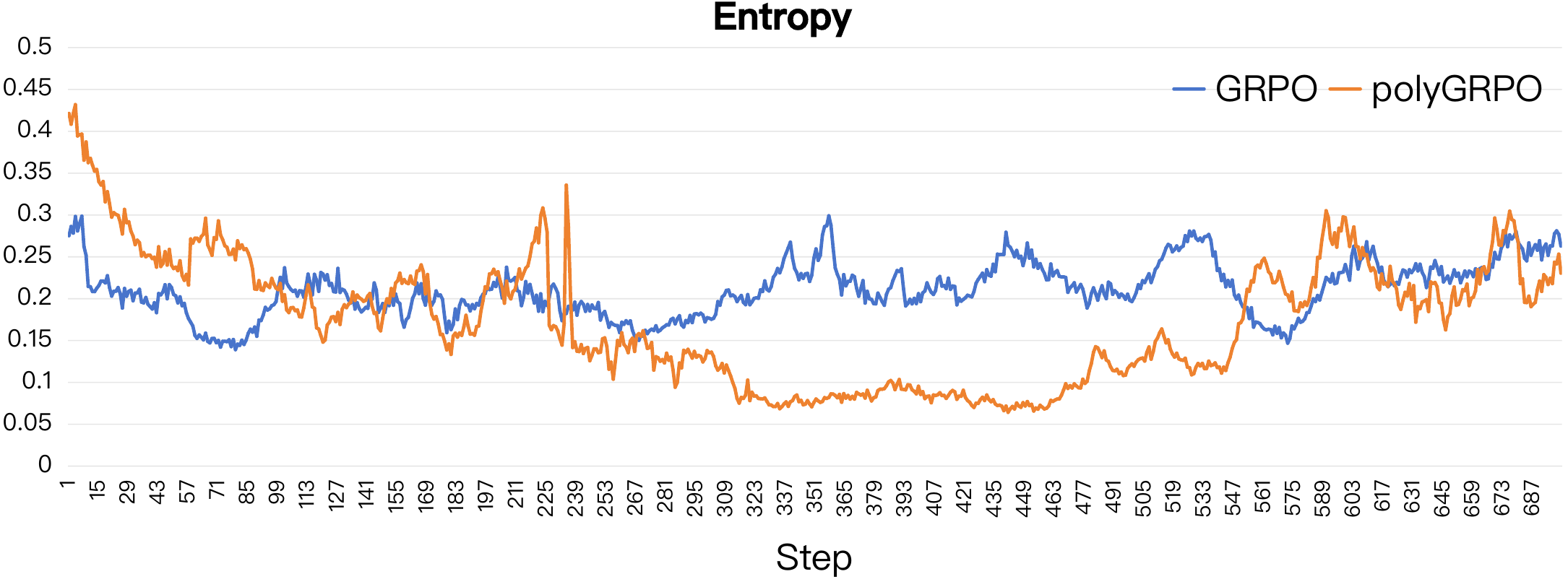}
    \caption{The entropy of GRPO and polyGRPO during training process based on Qwen2.5-7B-Instruct model with 10 epochs (700 steps total).}
    \label{entropy}
\end{figure}

We further analyze training dynamics via the entropy of the policy distribution over actions (token-level decisions). As shown in Figure \ref{entropy}, polyGRPO starts with significantly higher entropy than GRPO, reflecting greater stochasticity and exploration—likely attributable to the diverse Polyglot Thinking trajectories. Over time, the entropy of polyGRPO steadily decreases and stabilizes at a level lower than that of GRPO, indicating its policy becomes more confident and achieves optimal overall performance. The subsequent convergence of entropy to a level similar to GRPO is also intuitive, as polyGRPO increasingly relies on English for reasoning in later stages, aligning its training dynamics with those of GRPO.

Despite this eventual shift to English-dominated reasoning, a likely result of data imbalance or tokenization bias in base LLMs, the intermediate Polyglot Thinking phase is crucial in guiding optimization toward a more favorable solution.

\section{Conclusion}

This work introduces polyGRPO, a reinforcement learning framework that enhances reasoning in LLMs by treating language as a latent variable for reasoning optimization. By generating polyglot reasoning paths and optimizing rewards that account for both accuracy and reasoning format, polyGRPO encourages models to internalize diverse thinking strategies. Our results demonstrate that polyGRPO improves performance across four reasoning tasks in 23 languages, using both Qwen2.5 and Llama3 architectures. It achieves an average 6.4\% improvement over two base LLMs on MGSM, multilingual-version MATH500, and PolyMath-medium, while maintaining strong generalization to non-mathematical domains. Analysis shows that the model gradually shifts from multilingual to English reasoning during training, achieving better performance than training solely in English reasoning. This suggests that Polyglot Thinking traces act as scaffolding for stronger, language-agnostic reasoning capabilities. 

\section*{Impact Statement}
This work introduces polyGRPO, a reinforcement learning framework that treats polyglot thinking as a preference data to enhance LLM reasoning capabilities. 

Central to our work is the hypothesis that language functions not merely as an output medium, but as a latent structural modulator of reasoning. We validate this idea through polyGRPO, demonstrating consistent reasoning gains without requiring explicit chain-of-thought supervision. The approach is compatible with GRPO variants and other RL settings where preference signals are needed.
Beyond reasoning, the insight that language functions as a latent modulator of internal computation offers a general principle for optimizing diverse NLP tasks, such as machine translation, stylistic control, and even multi-modal reasoning, by harnessing cross-lingual inductive biases.

Nevertheless, important directions remain in this work. First, our current reward design relies only on final answers and coarse reasoning structures, overlooking fine-grained signals within the polyglot reasoning trajectories. Second, while we observe performance gains, we do not explicitly model the logical or structural relationships between reasoning paths in different languages, a dimension worth incorporating into future algorithm design. Third, evaluation focuses solely on answer correctness; reasoning process consistency of languages (e.g., the language of response alignment to question language) is a critical yet underexplored challenge for deploying LLMs in truly global user environments.

Differences in model behavior across languages reflect underlying reasoning strengths, which can be exploited via preference-based learning. We hope this motivates more work on turning multilingual variation into a structured signal for LLM training, rather than treating languages as independent channels.


\bibliography{example_paper}
\bibliographystyle{icml2026}

\newpage
\appendix
\onecolumn
\section{Benchmarks and Covered Languages}\label{bench_lang}

In this section, we provide the languages covered by the benchmarks used in our evaluation, as shown in Table~\ref{tab:datasets}.

\begin{table}[ht]
\centering
\small
\caption{Summary of benchmarks and their covered languages.}
\resizebox{1\textwidth}{!}{\begin{tabular}{lcl}
\toprule
\textbf{Dataset} & \textbf{\#Languages} & \textbf{Languages} \\
\midrule
MGSM~\cite{MGSM} & 10 & 
Chinese (ZH), French (FR), German (DE), Japanese (JA), Russian (RU), Spanish (ES) (HRL); \\
& & Bengali (BN), Swahili (SW), Telugu (TE), Thai (TH) (URL) \\

MATH500~\cite{MATH500} & 7 & 
English (EN), Chinese (ZH), Japanese (JA), Telugu (TE), Swahili (SW), Italian (IT), Turkish (TR) \\

PolyMath~\cite{wang2025polymath} & 18 & 
English (EN), Chinese (ZH), Spanish (ES), Arabic (AR), French (FR), Bengali (BN), Portuguese (PT), Russian (RU), \\
& & Indonesian (ID), German (DE), Japanese (JA), Swahili (SW), Vietnamese (VI), Italian (IT), Telugu (TE), Korean (KO), Thai (TH), Malay (MS) \\

X-CSQA~\cite{XCSQA} & 16 & 
Arabic (AR), German (DE), English (EN), Spanish (ES), French (FR), Hindi (HI), Italian (IT), Japanese (JA), \\
& & Dutch (NL), Polish (PL), Portuguese (PT), Russian (RU), Swahili (SW), Urdu (UR), Vietnamese (VI), Chinese (ZH) \\
\bottomrule
\end{tabular}}
\label{tab:datasets}
\end{table}

For MATH500 benchmark, the ZH, JA, TE, and SW versions of MATH500 are from https://huggingface.co/datasets/appier-ai-research; the IT and TR versions are from https://huggingface.co/datasets/bezir/MATH-500-multilingual.

\section{Baselines}\label{apx:baselines}
We compare polyGRPO with several strong baselines:
\begin{itemize}
\item The \textbf{base model}, Qwen2.5-7B-Instruct and Llama3-8B-Instruct, already demonstrates strong performance on reasoning tasks, which serves as a solid reference point.
\item \textbf{xRFT}~\cite{xrft} is a rejection sampling–based fine-tuning approach. It uses CoT traces generated by Qwen2.5-Math-7B-Instruct~\cite{Qwen-Math}, translated into multiple languages. After filtering for correctness and translation quality, around 9.7k multilingual CoT samples are retained. The model based on Qwen2.5-7B-Instruct is fine-tuned with a learning rate of 1e-5, batch size 128, for 3 epochs. Based on Llama3-8B-Instruct, the learning rate is 9e-7, batch size 64, for 1 epochs.
\item \textbf{LIDR} (Language Imbalance Driven Rewarding)~\cite{Self-improving} leverages performance gaps between dominant and underrepresented languages as implicit preference signals. LIDR applies DPO training on constructed preference bilingual CoT pairs in 10 languages align to our training data. Based on Qwen2.5-7B-Instruct and Llama3-8B-Instruct, we used 8.9K or 6.4k preference pairs data to train the LIDR model, respectively. The learning rate is 9e-7, batch size is 64, and epoch is 1 for both of them. 
\item \textbf{MAPO} (Multilingual-Alignment-as-Preference Optimization)~\cite{MAPO} aligns reasoning across languages using translation-based alignment scores. Following the original setup, we fine-tune using DPO with a learning rate of 1e-6, batch size 128, up to 1,000 steps, and select the best checkpoint based on validation loss.
\end{itemize}

\section{Per-language Results} \label{apx:perlang}
The per-language results on four benchmarks are shown in Table \ref{per-mgsm}, Table \ref{per-math500}, Table \ref{per-poly}, and Table \ref{per-xcsqa}.
\begin{table}[htbp]
  \centering
  \caption{The per-language results in MGSM benchmark.}
    \resizebox{0.8\textwidth}{!}{\begin{tabular}{lccc|c|cccccc|cccc}
    \toprule
    \textbf{MGSM} & AVG   & HRL   & URL   & EN    & DE    & FR    & ES    & RU    & ZH    & JA    & TH    & TE    & BN    & SW \\
    \midrule
    \rowcolor[rgb]{ .906,  .902,  .902} \textit{Qwen2.5-7B-Instruct} &       &       & \multicolumn{1}{r}{} & \multicolumn{1}{r}{} &       &       &       &       &       & \multicolumn{1}{r}{} &       &       &       &  \\
    Base  & 67.2  & 75.7  & 48.8  & 89.2  & 73.6  & 74.8  & 79.2  & 78.8  & 80.0  & 68.0  & 73.6  & 36.4  & 68.0  & 17.2  \\
    xRFT  & 68.5  & 81.1  & 43.2  & 94.0  & 81.6  & 78.0  & 84.4  & 83.6  & 85.2  & 73.6  & 69.6  & 25.2  & 54.4  & 23.6  \\
    LIDR  & 69.6  & 79.1  & 50.3  & 90.0  & 79.6  & 76.8  & 82.8  & 82.4  & 82.4  & 70.4  & 76.4  & 39.2  & 69.6  & 16.0  \\
    MAPO  & 66.3  & 75.8  & 47.4  & 84.8  & 78.4  & 76.4  & 79.2  & 78.8  & 74.8  & 67.2  & 74.4  & 33.2  & 62.8  & 19.2  \\
    GRPO  & 73.0  & 81.1  & 56.6  & 90.4  & 82.8  & 80.0  & 83.6  & 83.2  & 82.4  & 74.4  & 77.6  & 40.8  & 72.0  & 36.0  \\
    polyGRPO & \textbf{75.9} & \textbf{84.4} & \textbf{58.7} & 94.0  & 86.0  & 83.2  & 88.8  & 86.4  & 84.8  & 77.2  & 81.2  & 42.8  & 74.8  & 36.0  \\
    \midrule
    \rowcolor[rgb]{ .906,  .902,  .902} \textit{Llama3-8B-Instruct} &       &       & \multicolumn{1}{r}{} & \multicolumn{1}{r}{} &       &       &       &       &       & \multicolumn{1}{r}{} &       &       &       &  \\
    Base  & 52.2  & 57.3  & 37.7  & 79.6  & 59.6  & 60.8  & 63.6  & 59.2  & 57.6  & 43.2  & 48.4  & 26.0  & 41.2  & 35.2  \\
    xRFT  & 53.9  & 58.4  & 42.3  & 73.2  & 58.0  & 62.4  & 64.0  & 64.8  & 50.8  & 50.4  & 52.4  & 38.8  & 45.2  & 32.8  \\
    LIDR  & 55.5  & 58.5  & 45.1  & 79.6  & 66.4  & 62.4  & 61.6  & 61.2  & 52.8  & 46.4  & 56.4  & 43.6  & 35.2  & 45.2  \\
    MAPO  & 60.7  & 63.9  & 50.9  & 80.4  & 66.8  & 65.2  & 67.2  & 65.6  & 60.4  & 58.4  & 63.6  & 44.8  & 53.2  & 42.0  \\
    GRPO  & 64.6  & 68.8  & 54.2  & 80.8  & 72.0  & 72.4  & 72.4  & 70.0  & 64.4  & 61.6  & 62.8  & 48.4  & 57.6  & 48.0  \\
    polyGRPO & \textbf{68.1} & \textbf{72.3} & \textbf{58.3} & 82.0  & 74.8  & 75.2  & 80.0  & 74.8  & 65.6  & 63.6  & 63.2  & 51.2  & 64.8  & 54.0  \\
    \bottomrule
    \end{tabular}}%
  \label{per-mgsm}%
\end{table}%

\begin{table}[htbp]
  \centering
  \tiny
  \caption{The per-language results in MATH500 benchmark and its translation version in other 6 target languages.}
    \resizebox{0.7\textwidth}{!}{\begin{tabular}{lccc|c|cccc|cc}
    \toprule
    \textbf{MATH-500} & AVG   & HRL   & URL   & EN    & IT    & JA    & TR    & ZH    & TE    & SW \\
    \midrule
    \rowcolor[rgb]{ .906,  .902,  .902} \textit{Qwen2.5-7B-Instruct} &       &       & \multicolumn{1}{r}{} & \multicolumn{1}{r}{} &       &       &       & \multicolumn{1}{r}{} &       &  \\
    Base  & 57.9  & 60.9  & 45.5  & 70.8  & 68.4  & 61.6  & 51.8  & 61.6  & 52.4  & 38.6  \\
    xRFT  & 53.1  & 59.0  & 31.3  & 73.4  & 63.0  & 59.0  & 54.0  & 59.8  & 32.8  & 29.8  \\
    LIDR  & 62.5  & 66.7  & 48.6  & 73.2  & 70.6  & \textbf{68.4} & 65.6  & 62.2  & 57.2  & 40.0  \\
    MAPO  & 61.2  & 64.6  & 46.9  & 76.2  & 68.4  & 63.0  & 58.6  & 68.4  & 54.6  & 39.2  \\
    GRPO  & 63.2  & 68.0  & 47.8  & 74.8  & 69.6  & \textbf{68.4} & 68.4  & 65.6  & 56.2  & 39.4  \\
    polyGRPO & \textbf{64.9} & \textbf{69.8} & \textbf{49.2} & \textbf{76.8} & \textbf{71.8} & \textbf{68.4} & \textbf{70.4} & \textbf{68.8} & \textbf{57.6} & \textbf{40.8} \\
    \midrule
    \rowcolor[rgb]{ .906,  .902,  .902} \textit{Llama3-8B-Instruct} &       &       & \multicolumn{1}{r}{} & \multicolumn{1}{r}{} &       &       &       & \multicolumn{1}{r}{} &       &  \\
    Base  & 26.0  & 27.0  & 22.3  & 29.2  & 26.0  & 29.0  & 26.6  & \textbf{26.6}  & 25.4  & 19.2  \\
    xRFT  & 23.4  & 24.5  & 19.0  & 27.4  & 26.2  & 26.4  & 20.6  & 25.0  & 21.2  & 16.8  \\
    LIDR  & 24.3  & 24.5  & 22.1  & 28.0  & 25.8  & 25.4  & 24.2  & 22.6  & \textbf{26.0} & 18.2  \\
    MAPO  & 25.3  & 25.6  & 21.9  & 30.4  & 28.0  & 26.4  & 24.2  & 24.0  & 24.6  & 19.2  \\
    GRPO  & 24.4  & 25.7  & 19.1  & 30.0  & 28.0  & 24.6  & 24.8  & 25.4  & 19.2  & 19.0  \\
    polyGRPO & \textbf{26.7} & \textbf{27.1} & \textbf{23.2} & \textbf{32.0} & \textbf{29.8} & \textbf{27.2} & \textbf{26.4} & {25.2} & 24.6  & \textbf{21.8} \\
    \bottomrule
    \end{tabular}}%
  \label{per-math500}%
\end{table}%

\begin{table}[htbp]
  \centering
  \caption{The per-language results in four difficulty levels of PolyMath benchmark.}
    \resizebox{1\textwidth}{!}{\begin{tabular}{lc|cccccccccccccccccc}
    \toprule
    \multicolumn{1}{c}{\multirow{2}[2]{*}{Model}} & \multicolumn{19}{c}{PolyMath-Medium} \\
          & AVG  & EN    & ZH    & ES    & AR    & FR    & BN    & PT    & RU    & ID    & DE    & JA    & SW    & VI    & IT    & TE    & KO    & TH    & MS \\
    \midrule
    \rowcolor[rgb]{ .906,  .902,  .902} \textit{Qwen2.5-7B-Instruct} & & & & & & & & & & & & & & & & & & &\\
    Base & 24.00  & 26.4  & 20.0  & 24.8  & 21.6  & 29.6  & 25.6  & 23.2  & 27.2  & 26.4  & 27.2  & 20.8  & 14.4  & 24.8  & 26.4  & 20.8  & 25.6  & 18.4  & 23.2  \\
    xRFT  & 19.20  & 28.8  & 16.0  & 21.6  & 20.0  & 21.6  & 15.2  & 20.0  & 20.8  & 19.2  & 16.8  & 19.2  & 12.8  & 17.6  & 21.6  & 16.0  & 13.6  & 14.4  & 16.0  \\
    LIDR  & 25.05  & 28.0  & 20.0  & 28.8  & 26.4  & 27.2  & 23.2  & 27.2  & 26.4  & 27.2  & 24.0  & 22.4  & 18.4  & 26.4  & 29.6  & 16.8  & 25.6  & 28.0  & 25.6  \\
    MAPO  & 23.02  & 32.0  & 23.2  & 24.8  & 23.2  & 22.4  & 2.4   & 21.6  & 29.6  & 23.2  & 27.2  & 24.0  & 17.6  & 28.0  & 25.6  & 24.0  & 20.8  & 23.2  & 25.6  \\
    GRPO  & 23.88  & 26.4  & 23.2  & 26.4  & 24.0  & 22.4  & 18.4  & 25.6  & 28.8  & 24.8  & 24.0  & 22.4  & 17.6  & 26.4  & 27.2  & 18.4  & 24.0  & 21.6  & 20.0  \\
    polyGRPO & 25.97  & 29.6  & 31.2  & 26.4  & 31.2  & 24.8  & 22.4  & 28.0  & 26.4  & 24.0  & 25.6  & 23.2  & 21.6  & 23.2  & 27.2  & 20.0  & 26.4  & 22.4  & 20.8  \\
    \midrule
    \rowcolor[rgb]{ .906,  .902,  .902} \textit{Llama3-8B-Instruct} & & & & & & & & & & & & & & & & & & &\\
    Base & 4.49  & 8.0   & 3.2   & 6.4   & 5.6   & 2.4   & 2.4   & 4.8   & 5.6   & 5.6   & 4.0   & 6.4   & 2.4   & 1.6   & 4.0   & 4.8   & 7.2   & 6.4   & 6.4  \\
    xRFT  & 5.17  & 6.4   & 8.0   & 7.2   & 4.8   & 3.2   & 2.4   & 5.6   & 5.6   & 4.0   & 5.6   & 3.2   & 3.2   & 8.0   & 7.2   & 0.8   & 4.8   & 3.2   & 5.6  \\
    LIDR  & 4.98  & 7.2   & 7.2   & 4.0   & 4.8   & 5.6   & 1.6   & 7.2   & 3.2   & 2.4   & 7.2   & 6.4   & 4.0   & 4.0   & 9.6   & 3.2   & 5.6   & 3.2   & 4.0  \\
    MAPO  & 4.68  & 6.4   & 2.4   & 4.0   & 4.8   & 7.2   & 5.6   & 4.8   & 4.8   & 4.0   & 4.8   & 5.6   & 2.4   & 4.0   & 0.8   & 4.8   & 3.2   & 4.8   & 3.2  \\
    GRPO  & 5.11  & 8.0   & 2.4   & 7.2   & 5.6   & 6.4   & 2.4   & 6.4   & 2.4   & 8.0   & 4.8   & 3.2   & 3.2   & 6.4   & 4.8   & 4.0   & 4.0   & 3.2   & 3.2  \\
    polyGRPO & 5.54  & 9.6   & 4.0   & 8.8   & 4.8   & 4.0   & 4.0   & 6.4   & 4.0   & 5.6   & 7.2   & 4.0   & 4.0   & 5.6   & 6.4   & 6.4   & 5.6   & 4.0   & 4.8  \\
    \bottomrule
    \end{tabular}}%
  \label{per-poly}%
\end{table}%

\begin{table}[htbp]
  \centering
  \caption{The per-language results in X-CSQA benchmark.}
    \resizebox{1\textwidth}{!}{\begin{tabular}{l|c|cccccccccccccccc}
    \toprule
    \textbf{Model} & AVG   & AR    & DE    & EN    & ES    & FR    & HI    & IT    & JA    & NL    & PL    & PT    & RU    & SW    & UR    & VI    & ZH \\
    \midrule
    \rowcolor[rgb]{ .906,  .902,  .902} \multicolumn{1}{l}{\textit{Qwen2.5-7B-Instruct}} & \multicolumn{1}{r}{} &       &       &       &       &       &       &       &       &       &       &       &       &       &       &       &  \\
    Base  & 54.3  & 53.0  & 56.0  & 77.1  & 62.9  & 59.3  & 43.2  & 59.5  & 50.6  & 58.0  & 53.8  & 63.6  & 52.9  & 25.5  & 36.3  & 56.0  & 60.6  \\
    xRFT  & 49.3  & 46.9  & 60.5  & 70.3  & 61.0  & 56.8  & 37.2  & 56.1  & 43.4  & 55.1  & 53.4  & 60.3  & 39.3  & 17.3  & 27.3  & 50.1  & 53.0  \\
    LIDR  & 53.2  & 53.3  & 58.1  & 75.1  & 60.5  & 57.0  & 40.7  & 54.6  & 52.5  & 54.1  & 55.1  & 56.0  & 54.1  & 26.3  & 36.2  & 57.3  & 60.0  \\
    MAPO  & 50.7  & 48.6  & 49.7  & 77.1  & 61.1  & 55.5  & 39.2  & 55.2  & 45.2  & 51.9  & 49.8  & 54.7  & 50.7  & 25.1  & 33.4  & 56.5  & 57.3  \\
    GRPO  & 57.1  & 56.4  & 62.4  & 75.5  & 64.1  & 62.1  & 46.1  & 62.5  & 57.1  & 59.8  & 60.8  & 62.6  & 58.2  & 28.3  & 37.8  & 60.1  & 59.8  \\
    polyGRPO (Ours) & \textbf{60.5} & \textbf{58.8} & \textbf{65.1} & \textbf{82.0} & \textbf{68.4} & \textbf{67.0} & \textbf{50.3} & \textbf{66.8} & \textbf{57.7} & \textbf{61.7} & \textbf{62.3} & \textbf{65.9} & \textbf{62.0} & \textbf{31.2} & \textbf{40.3} & \textbf{63.5} & \textbf{64.4} \\
    \midrule
    \rowcolor[rgb]{ .906,  .902,  .902} \multicolumn{1}{l}{\textit{Llama3-8B-Instruct}} & \multicolumn{1}{r}{} &       &       &       &       &       &       &       &       &       &       &       &       &       &       &       &  \\
    Base  & 45.1  & 41.7  & 49.7  & 66.3  & 50.6  & 50.6  & 36.8  & 48.1  & 39.0  & 46.6  & 42.0  & 49.3  & 45.7  & 31.4  & 32.5  & 45.8  & 45.8  \\
    xRFT  & 48.2  & 46.2  & 51.8  & 67.7  & 54.9  & 53.3  & 40.7  & 50.4  & 42.2  & 49.7  & 46.4  & 52.8  & 50.4  & 33.1  & 35.2  & 48.3  & 47.3  \\
    LIDR  & 52.2  & 47.0  & 55.5  & 69.5  & 57.6  & 55.9  & 46.9  & 55.4  & 47.6  & 52.8  & 51.5  & 55.3  & 54.8  & 38.5  & 41.4  & 52.4  & 53.4  \\
    MAPO  & 43.9  & 42.4  & 48.9  & 62.3  & 48.1  & 44.7  & 37.5  & 46.6  & 39.0  & 47.0  & 42.1  & 48.8  & 42.7  & 27.8  & 33.2  & 46.0  & 45.2  \\
    GRPO  & 53.4  & 50.4  & 57.0  & 68.8  & 59.3  & 57.6  & 45.4  & 55.5  & 49.1  & \textbf{56.4} & \textbf{53.7} & \textbf{59.2} & 53.8  & \textbf{40.0} & 40.2  & 52.0  & \textbf{55.4} \\
    polyGRPO (Ours) & \textbf{53.6} & \textbf{50.9} & \textbf{57.6} & \textbf{70.9} & \textbf{60.2} & \textbf{58.2} & \textbf{47.0} & \textbf{57.1} & \textbf{50.1} & 55.0  & 52.1  & 57.1  & \textbf{54.1} & 39.1  & \textbf{42.1} & \textbf{54.5} & 52.3  \\
    \bottomrule
    \end{tabular}}%
  \label{per-xcsqa}%
\end{table}%

\section{Ablation Study}\label{apx:ablation}
The details results of our ablation study is shown in Table~\ref{apx:abl}.
\begin{table}[htbp]
  \centering
  \caption{The results of Ablation Study on MGSM. Best in \textbf{bold}.}
    \resizebox{0.8\textwidth}{!}{\begin{tabular}{r|ccc|c|cccccc|cccc}
    \toprule
    \multicolumn{1}{l|}{\textbf{MGSM}} & AVG   & HRL   & URL   & EN    & DE    & FR    & ES    & RU    & ZH    & JA    & TH    & TE    & BN    & SW \\
    \midrule
    \multicolumn{1}{l|}{Qwen2.5-7B-Instruct} & 67.2  & 75.7  & 48.8  & 89.2  & 73.6  & 74.8  & 79.2  & 78.8  & 80.0  & 68.0  & 73.6  & 36.4  & 68.0  & 17.2  \\
    \multicolumn{1}{l|}{GRPO}   & 73.0  & 81.1  & 56.6  & 90.4  & 82.8  & 80.0  & 83.6  & 83.2  & 82.4  & 74.4  & 77.6  & 40.8  & 72.0  & \textbf{36.0} \\
    \multicolumn{1}{l|}{polyGRPO} & \textbf{75.9} & \textbf{84.4} & \textbf{58.7} & \textbf{94.0} & \textbf{86.0} & \textbf{83.2} & \textbf{88.8} & \textbf{86.4} & \textbf{84.8} & 77.2  & \textbf{81.2} & 42.8  & \textbf{74.8} & \textbf{36.0} \\
    w/o format reward & 71.7  & 79.7  & 55.4  & 88.4  & 78.8  & 79.6  & 83.6  & 82.0  & 79.2  & 75.2  & 80.8  & 41.2  & 66.0  & 33.6  \\
    w/o unconstrained response   & 75.6  & \textbf{84.4} & 58.0  & 92.8  & 84.0  & 81.6  & 86.8  & 85.6  & \textbf{85.6} & \textbf{82.8} & \textbf{81.2} & 43.2  & 74.4  & 33.2  \\
    only roll-out unconstrained response   & 70.4  & 78.2  & 54.7  & 86.4  & 79.2  & 77.2  & 80.4  & 80.4  & 76.4  & 75.6  & 79.6  & \textbf{43.6} & 64.4  & 31.2  \\
    \midrule
    \multicolumn{1}{l|}{Qwen2.5-3B-Instruct} & 52.4  & 63.1  & 30.4  & 76.4  & 64.4  & 66.4  & 65.6  & 62.4  & 64.0  & 56.0  & 56.0  & 14.4  & 40.4  & 10.8  \\
    \multicolumn{1}{l|}{GRPO}   & 61.7  & 72.6  & 39.7  & 84.4  & 74.0  & 73.6  & 79.2  & 71.6  & 73.2  & 64.0  & 68.8  & 21.2  & 53.2  & 15.6  \\
    \multicolumn{1}{l|}{polyGRPO} & 64.7  & 74.9  & \textbf{44.3} & 84.4  & 75.6  & 76.8  & \textbf{80.0} & 75.6  & 74.8  & \textbf{66.8} & \textbf{75.2} & \textbf{23.6} & \textbf{62.4} & 16.0  \\
    w/o format reward & 61.6  & 71.3  & 42.1  & 82.0  & 76.4  & 69.2  & 73.2  & 73.6  & 70.8  & 64.4  & 65.2  & 23.2  & 60.0  & 20.0  \\
    w/o unconstrained response   & 59.3  & 69.2  & 38.8  & 81.6  & 73.2  & 72.0  & 71.2  & 70.0  & 69.6  & 59.2  & 64.0  & 22.4  & 54.4  & 14.4  \\
    only roll-out unconstrained response   & \textbf{64.9} & \textbf{76.0} & 42.7  & \textbf{86.8} & \textbf{78.8} & \textbf{77.6} & 79.6  & \textbf{78.4} & \textbf{76.0} & 65.6  & 70.8  & 21.6  & 57.6  & \textbf{20.8} \\
    \midrule
    \multicolumn{1}{l|}{Qwen2.5-1.5B-Instruct} & 26.1  & 31.5  & 14.1  & 41.2  & 24.4  & 29.6  & 34.4  & 35.2  & 40.0  & 25.6  & 29.2  & 6.4   & 17.6  & 3.2  \\
    \multicolumn{1}{l|}{GRPO}   & 47.2  & 58.9  & 23.5  & 72.0  & \textbf{63.6} & 60.0  & 62.0  & 62.8  & 58.4  & 46.4  & 47.2  & 8.8   & 32.0  & 6.0  \\
    \multicolumn{1}{l|}{polyGRPO} & \textbf{51.0} & \textbf{62.5} & \textbf{26.9} & \textbf{78.4} & 61.6  & \textbf{65.6} & \textbf{66.8} & \textbf{66.0} & \textbf{65.6} & \textbf{49.6} & \textbf{53.6} & \textbf{12.8} & \textbf{34.0} & 7.2  \\
    w/o format reward & 14.0  & 16.9  & 7.5   & 23.2  & 17.2  & 15.2  & 15.2  & 16.4  & 24.0  & 13.2  & 12.4  & 4.8   & 8.0   & 4.8  \\
    w/o unconstrained response   & 47.5  & 57.7  & 25.1  & 76.0  & 58.4  & 58.8  & 64.8  & 60.8  & 56.4  & 46.8  & 50.0  & 11.6  & 28.4  & \textbf{10.4} \\
    only roll-out unconstrained response   & 47.9  & 59.7  & 23.7  & 73.6  & 59.2  & 62.0  & 64.0  & 62.4  & 62.0  & 48.4  & 47.6  & 10.4  & 29.2  & 7.6  \\
    \bottomrule
    \end{tabular}}%
  \label{apx:abl}%
\end{table}%

\section{The number of Language Sets} \label{apx:l_set_num}
To assess the effect of language diversity, we expand it to 15 by adding Arabic (AR), Korean (KO), Portuguese (PT), Telugu (TE), and Vietnamese (VI). We also evaluate reduced settings by randomly selecting 5 languages from the original set, repeating this process three times to assess stability. All experiments are conducted on Qwen2.5-1.5B-Instruct and evaluated on MGSM. Results are shown in Table~\ref{roll-out_num}, the 15-language setup slightly hurts overall performance. Reducing to 5 languages leads to further degradation and high variance depending on language selection. These findings indicate that the original 10-language configuration offers a good trade-off between diversity and stability of language sets.

\begin{table}[ht]
  \centering
  \caption{Effects of different number of language sets on MGSM with Qwen2.5-1.5B-Instruct.}
    \resizebox{1\textwidth}{!}{\begin{tabular}{r|ccc|c|cccccc|cccc}
    \toprule
    \multicolumn{1}{l|}{\textbf{MGSM}} & AVG   & HRL   & URL   & EN    & DE    & FR    & ES    & RU    & ZH    & JA    & TH    & TE    & BN    & SW \\
    \midrule
    \multicolumn{1}{l|}{Qwen2.5-1.5B-Instruct} & 26.1  & 31.5  & 14.1  & 41.2  & 24.4  & 29.6  & 34.4  & 35.2  & 40.0  & 25.6  & 29.2  & 6.4   & 17.6  & 3.2  \\
    \multicolumn{1}{l|}{polyGRPO} & \textbf{51.0} & \textbf{62.5} & \textbf{26.9} & \textbf{78.4} & \textbf{61.6} & \textbf{65.6} & \textbf{66.8} & \textbf{66.0} & \textbf{65.6} & \textbf{49.6} & 53.6  & \textbf{12.8} & \textbf{34.0} & \textbf{7.2} \\
    \text{  }lang\_num=15 & 48.7  & 60.5  & 24.7  & 73.2  & 59.6  & 64.0  & 64.4  & 64.0  & 64.4  & 46.8  & \textbf{54.0} & 9.6   & 28.4  & 6.8  \\
    \text{\ \ \ \ \ }lang\_num=5, (DE, EN, ES, RU, SW) & 47.6  & 59.0  & 24.8  & 70.0  & 59.2  & 62.0  & 61.6  & 58.4  & 63.2  & \textbf{49.6} & 50.0  & 8.4   & 33.6  & \textbf{7.2} \\
    \text{\ \ \ \ \ }lang\_num=5, (ES, FR, SW, TH, ZH) & 44.0  & 54.2  & 22.6  & 68.4  & 56.0  & 55.2  & 58.4  & 57.2  & 54.8  & 43.6  & 44.8  & 10.4  & 29.6  & 5.6  \\
    \text{\ \ \ \ \ }lang\_num=5, (ES, FR, RU, SW, ZH) & 15.6  & 20.1  & 6.6   & 25.2  & 18.0  & 20.8  & 16.0  & 17.6  & 30.4  & 17.6  & 10.0  & 2.4   & 7.6   & 6.4  \\
    \bottomrule
    \end{tabular}}%
  \label{roll-out_num}%
\end{table}%

\section{Roll-out Number} \label{apx:n}
We also study the effect of varying the roll-out number $n \in \{4, 8, 10, 16\}$ in 1.5B model, and results shown in the top of Table~\ref{tab:roll4}. The best performance is observed with $n = 4$. Increasing $n$ to 8 already causes a noticeable drop in performance then ours. With $n = 10$, training becomes unstable due to overexposure to low-resource languages, and we observe a significant amount of garbled text in the model's outputs during later training stages. At $n = 16$, duplicate sampling mitigates some instability. To test whether $n = 4$ generalizes to other model sizes, we also evaluate $n = 4$ on the 3B and 7B models in the Table~\ref{tab:roll4} and lower than $n=5$.

\begin{table}[htbp]
  \centering
  \caption{Performance comparison when using different roll-out number (n) in our polyGRPO based on three Qwen2.5-Instruct models.}
    \resizebox{0.8\textwidth}{!}{\begin{tabular}{l|ccc|c|cccccc|cccc}
    \toprule
    \multicolumn{1}{l|}{\textbf{MGSM}} & AVG   & HRL   & URL   & EN    & DE    & FR    & ES    & RU    & ZH    & JA    & TH    & TE    & BN    & SW \\
    \midrule
    \midrule
    \multicolumn{1}{l|}{\textbf{Qwen2.5-1.5B-Instruct}} & 26.1  & 31.5  & 14.1  & 41.2  & 24.4  & 29.6  & 34.4  & 35.2  & 40.0  & 25.6  & 29.2  & 6.4   & 17.6  & 3.2  \\
    \midrule
    n=4   & \textbf{52.0} & 62.3  & \textbf{29.3} & \textbf{80.8} & 60.8  & 64.8  & \textbf{68.0} & 60.8  & \textbf{67.2} & \textbf{52.4} & \textbf{56.4} & \textbf{16.0} & \textbf{35.6} & \textbf{9.2} \\
    \multicolumn{1}{l|}{n=5} & 51.0  & \textbf{62.5} & 26.9  & 78.4  & \textbf{61.6} & \textbf{65.6} & 66.8  & \textbf{66.0} & 65.6  & 49.6  & 53.6  & 12.8  & 34.0  & 7.2  \\
    n=8   & 48.3  & 59.3  & 24.8  & 76.8  & 60.8  & 60.4  & 62.4  & 62.0  & 63.2  & 46.8  & 47.6  & 9.2   & 34.0  & 8.4  \\
    n=10  & 15.2  & 19.1  & 8.5   & 18.8  & 17.6  & 17.2  & 18.8  & 20.4  & 25.2  & 15.6  & 16.0  & 5.2   & 7.6   & 5.2  \\
    n=16  & 39.9  & 49.9  & 19.3  & 62.4  & 48.4  & 53.2  & 50.8  & 53.6  & 50.8  & 42.8  & 38.0  & 8.4   & 23.6  & 7.2  \\
     \midrule \midrule
    \textbf{Qwen2.5-3B-Instruct} & AVG   & HRL   & URL   & EN    & DE    & FR    & ES    & RU    & ZH    & JA    & TH    & TE    & BN    & SW \\
    \midrule
    n=5   & \textbf{64.7} & \textbf{74.9} & \textbf{44.3} & \textbf{84.4} & \textbf{75.6} & \textbf{76.8} & \textbf{80.0} & \textbf{75.6} & \textbf{74.8} & \textbf{66.8} & \textbf{75.2} & \textbf{23.6} & \textbf{62.4} & \textbf{16.0} \\
    n=4   & 59.4  & 69.5  & 38.8  & 81.2  & \textbf{75.6} & 72.0  & 70.4  & 70.8  & 68.0  & 60.4  & 67.2  & 19.6  & 54.0  & 14.4  \\
    \midrule
    \midrule
    \textbf{Qwen2.5-7B-Instruct} & AVG   & HRL   & URL   & EN    & DE    & FR    & ES    & RU    & ZH    & JA    & TH    & TE    & BN    & SW \\
    \midrule
    n=5   & \textbf{75.9} & \textbf{84.4} & \textbf{58.7} & \textbf{94.0} & \textbf{86.0} & \textbf{83.2} & \textbf{88.8} & \textbf{86.4} & \textbf{84.8} & 77.2  & \textbf{81.2} & \textbf{42.8} & \textbf{74.8} & \textbf{36.0} \\
    n=4   & 73.5  & 82.1  & 56.3  & 90.8  & 84.0  & 81.2  & 85.6  & 83.6  & 80.4  & \textbf{78.0} & 79.2  & 39.2  & 72.0  & 34.8  \\
    \bottomrule
    \end{tabular}}%
  \label{tab:roll4}%
\end{table}%

\section{The performance on Base LLM}
Since LIDR and MAPO are evaluated on Instruct models, our main experiments use Instruct models as well. We also experimented on Qwen2.5-7B base models. However, we observed that base models indeed lack strong instruction-following capabilities, often leading to undesirable continuation behaviors, such as generating additional samples (e.g., "\#\#\# Instruction:" right after "\#\#\#\# final answer").

\begin{table}[htbp]
  \centering
  \caption{The results in MGSM benchmark based on two base LLM, Qwen2.5-7B and Qwen3-8B.}
    \resizebox{1\textwidth}{!}{\begin{tabular}{lccc|c|cccccc|cccc}
    \toprule
    \textbf{Model} & AVG   & HRL   & URL   & EN    & DE    & FR    & ES    & RU    & ZH    & JA    & TH    & TE    & BN    & SW \\
    \midrule
    \rowcolor[rgb]{ .906,  .902,  .902} \textit{Qwen2.5-7B} &       &       & \multicolumn{1}{r}{} & \multicolumn{1}{r}{} &       &       &       &       &       & \multicolumn{1}{r}{} &       &       &       &  \\
    Base  & 51.45  & 65.27  & 25.00  & 74.40  & 64.0  & 63.2  & 68.8  & 69.2  & 73.2  & 53.2  & 38.8  & 13.2  & 38.8  & 9.2  \\
    LIDR  & 61.67  & 70.33  & 41.90  & 88.80  & 67.6  & 68.4  & 69.6  & 72.4  & 79.2  & 64.8  & 68.0  & 24.0  & 55.6  & 20.0  \\
    MAPO  & 61.85  & 72.20  & 40.30  & 86.00  & 68.0  & 71.6  & 79.6  & 76.4  & 75.6  & 62.0  & 65.2  & 24.8  & 51.6  & 19.6  \\
    GRPO  & 70.98  & 78.67  & 54.60  & 90.40  & 80.4  & 79.6  & 80.8  & 81.2  & 80.4  & 69.6  & 77.6  & 39.6  & 65.2  & 36.0  \\
    polyGRPO & 74.76  & 82.53  & 58.80  & \textbf{92.00} & 82.0  & 82.8  & \textbf{85.6} & \textbf{87.2} & 84.4  & 73.2  & 83.6  & 42.8  & 70.8  & \textbf{38.0} \\
    polyGRPO w/ R1-format & \textbf{76.07} & \textbf{83.80} & \textbf{61.30} & 88.80  & \textbf{82.8} & \textbf{83.2} & 83.6  & 86.0  & \textbf{88.0} & \textbf{79.2} & \textbf{84.8} & \textbf{46.8} & \textbf{75.6} & \textbf{38.0} \\
    \midrule
    \rowcolor[rgb]{ .906,  .902,  .902} \textit{Qwen3-8B} &       &       & \multicolumn{1}{r}{} & \multicolumn{1}{r}{} &       &       &       &       &       & \multicolumn{1}{r}{} &       &       &       &  \\
    Base  & 81.45  & 84.93  & 73.20  & 93.60  & 84.4  & 82.0  & 86.8  & 88.0  & 85.2  & 83.2  & 84.4  & 72.8  & 80.0  & 55.6  \\
    polyGRPO w/ R1-format & \textbf{87.27} & \textbf{90.47} & \textbf{80.00} & \textbf{97.20} & \textbf{91.2} & \textbf{90.8} & \textbf{92.4} & \textbf{94.4} & \textbf{88.4} & \textbf{85.6} & \textbf{90.8} & \textbf{79.2} & \textbf{88.8} & \textbf{61.2} \\
    \bottomrule
    \end{tabular}}%
  \label{base_results}%
\end{table}%

To address this, we introduced a penalty term in polyGRPO's format reward: if the model generates such continuations, we subtract 0.5 from the original format reward. For LIDR and MAPO, we directly removed the continuation content during preference data preparation.  The experimental results are shown in the top part of Table~\ref{base_results}. polyGRPO also obtain the SOTA score in MGSM benchmark trained on the base LLM.

We also conducted polyGRPO training on newest Qwen3-8B, directly adopting the R1 format used in its original training. The R1 format is to places the reasoning process within "<think>...<\textbackslash think>" tags and sets the output format to "\textbackslash boxed\{final answer\}".  The model trained with R1 format is named polyGRPO w/ R1 format. The results are shown in bottom of Table~\ref{base_results} and represents that polyGRPO is suited for R1 format output and even obtain better performance; and for stronger base LLM, Qwen3-8B, polyGRPO also could obtain improvement of performance.

\section{How to generate a response language consistent with the query language?}\label{end2end}
Our method focuses on enhancing reasoning capabilities through Polyglot Thinking, which has shown promising results on both mathematical and commonsense reasoning benchmarks. However, the reasoning process itself predominantly converges toward English, even when the questions are in other languages. This reliance on English limits the direct applicability of the model in user-facing multilingual scenarios.

\textbf{We explore a simple test-time strategy that enables polyGRPO to achieve a significantly higher language accuracy with only a minor performance trade-off.} We experiment with prepending language-specific prefixes (e.g., “Okay,” for English, “D'accord,” for French, “Sawa,” for Swahili) after the input to guide the model reason in user language. The user language is identified with langid Tool. Besides accuracy, we add a language consistency score to measure whether the generated reasoning matches the query language. The results in MGSM is shown in Table~\ref{apx_lang1}. With these prefix, polyGRPO obtain 100\% language consistency in 10 languages expect low-resource Swahili, while still outperforming GRPO.

\begin{table}[htbp]
  \centering
  \caption{The results of \textbf{Accuracy} and \textbf{Language Consistency} on MGSM with language control by language-specific prefix during inference.}
    \resizebox{1\textwidth}{!}{\begin{tabular}{rccc|c|cccccc|cccc}
    \toprule
    \multicolumn{1}{c}{\multirow{2}[4]{*}{\textbf{Model}}} & \multicolumn{14}{c}{Accuracy} \\
\cmidrule{2-15}          & AVG   & HRL   & URL   & EN    & DE    & FR    & ES    & RU    & ZH    & JA    & TH    & TE    & BN    & SW \\
    \midrule
    \multicolumn{1}{l}{Qwen2.5-7B-Instruct} & 67.2  & 75.7  & 48.8  & 89.2  & 73.6  & 74.8  & 79.2  & 78.8  & 80.0  & 68.0  & 73.6  & 36.4  & 68.0  & 17.2  \\
    w/ prefix & 66.8  & 76.5  & 46.4  & 90.0  & 75.2  & 74.8  & 81.2  & 79.6  & 80.8  & 67.2  & 74.4  & 33.6  & 61.2  & 16.4  \\
    \multicolumn{1}{l}{GRPO} & 73.0  & 81.1  & 56.6  & 90.4  & 82.8  & 80.0  & 83.6  & 83.2  & 82.4  & 74.4  & 77.6  & 40.8  & 72.0  & 36.0  \\
    w/ prefix & 72.4  & 82.4  & 52.7  & 91.6  & 83.6  & 78.8  & 86.4  & 85.2  & \textbf{85.2} & 75.2  & 80.4  & 35.2  & 70.8  & 24.4  \\
    \multicolumn{1}{l}{polyGRPO} & \textbf{75.9} & \textbf{84.4} & \textbf{58.7} & \textbf{94.0} & \textbf{86.0} & \textbf{83.2} & \textbf{88.8} & \textbf{86.4} & 84.8  & \textbf{77.2} & \textbf{81.2} & \textbf{42.8} & \textbf{74.8} & \textbf{36.0} \\
    w/ prefix & 74.3  & 83.7  & 55.7  & 92.4  & 83.6  & \textbf{83.2} & 88.0  & 85.2  & \textbf{85.2} & 76.8  & 79.2  & 38.4  & 70.8  & 34.4  \\
    \midrule
    \multicolumn{1}{c}{\multirow{2}[4]{*}{\textbf{Model}}} & \multicolumn{14}{c}{Language Consistency} \\
\cmidrule{2-15}          & AVG   & HRL   & URL   & EN    & DE    & FR    & ES    & RU    & ZH    & JA    & TH    & TE    & BN    & SW \\
    \midrule
    \multicolumn{1}{l}{Qwen2.5-7B-Instruct} & 68.4  & 92.6  & 24.3  & 100.0  & 80.0  & 94.8  & 98.0  & 95.2  & 100.0  & 87.6  & 38.8  & 16.0  & 2.4   & 40.0  \\
    w/ prefix & 99.7  & 99.9  & 99.6  & 99.6  & 100.0  & 100.0  & 100.0  & 99.2  & 100.0  & 100.0  & 99.2  & 99.6  & 99.6  & 100.0  \\
    \multicolumn{1}{l}{GRPO} & 17.8  & 14.4  & 2.3   & 100.0  & 1.6   & 3.2   & 1.6   & 0.4   & 74.4  & 5.2   & 1.2   & 0.4   & 1.2   & 6.4  \\
    w/ prefix & 99.7  & 99.9  & 99.4  & 100.0  & 99.6  & 100.0  & 100.0  & 100.0  & 100.0  & 100.0  & 100.0  & 100.0  & 99.6  & 98.0  \\
    \multicolumn{1}{l}{polyGRPO} & 9.6   & 0.2   & 1.1   & 99.6  & 0.0   & 0.4   & 0.0   & 0.0   & 0.4   & 0.4   & 0.4   & 2.4   & 0.0   & 1.6  \\
    w/ prefix & 99.1  & 100.0  & 97.4  & 100.0  & 100.0  & 100.0  & 100.0  & 100.0  & 100.0  & 100.0  & 100.0  & 100.0  & 100.0  & 89.6  \\
    \bottomrule
    \end{tabular}}%
  \label{apx_lang1}%
\end{table}%

To enable the model to perform reasoning in the input language, we also attempt a new version of polyGRPO, named \textbf{polyGRPO$_{lang}$}. This version introduces two main modifications: first, all prompts in the PRGM module are constrained response-language; second, a language consistency reward is added to the reward module as a language control signal, as mentioned in GRPO~\citep{r1}. We use the FastText~\citep{fasttext,fasttext1} to detect the language of the generated reasoning. When the generated language matches the prompt language, the reward is set to 1; otherwise, it is 0. We train polyGRPO$_{lang}$ on the Qwen2.5-7B-Instruct model, keeping all other training parameters the same as before. Evaluation is conducted mainly on the MGSM dataset, with both unconstrained and language-constrained prompts. The results, shown in the Table~\ref{apx_lang}, although polyGRPO$_{lang}$ achieves better language consistency across most languages, it experiments a drop in accuracy, especially in low-resource languages.

\begin{table}[htbp]
  \centering
  \caption{The results of \textbf{Accuracy} on MGSM and the \textbf{Language Consistency} between query and response languages.}
    \resizebox{1\textwidth}{!}{\begin{tabular}{lccc|c|cccccc|cccc}
    \toprule
    \multicolumn{1}{c}{\multirow{2}[4]{*}{\textbf{Model}}} & \multicolumn{14}{c}{Accuracy} \\
\cmidrule{2-15}          & AVG   & HRL   & URL   & EN    & DE    & FR    & ES    & RU    & ZH    & JA    & TH    & TE    & BN    & SW \\
    \midrule
    Qwen2.5-7B-Instruct & 67.2  & 75.7  & 48.8  & 89.2  & 73.6  & 74.8  & 79.2  & 78.8  & 80.0  & 68.0  & 73.6  & 36.4  & 68.0  & 17.2  \\
    polyGRPO & \textbf{75.9} & \textbf{84.4} & \textbf{58.7} & \textbf{94.0} & {86.0} & \textbf{83.2} & \textbf{88.8} & \textbf{86.4} & {84.8} & \textbf{77.2} & \textbf{81.2} & \textbf{42.8} & \textbf{74.8} & {36.0} \\
    polyGRPO$_{lang}$ w/ unconstrained prompt & 74.4 	& 83.4 & 56.0 & 93.6  & \textbf{86.8}  & 81.2  & 88.4  & 85.2  & \textbf{85.6}  & 73.2  & 79.6  & 44.8  & 72.8  & \textbf{26.8}  \\
    polyGRPO$_{lang}$ w/ language-constrained prompt & 66.0  & 78.1  & 42.3  & 87.6  & 78.8  & 78.0  & 80.8  & 81.2  & 76.0  & 74.0  & 77.2  & 12.8  & 61.6  & 17.6  \\
    \midrule
    \multicolumn{1}{c}{\multirow{2}[4]{*}{\textbf{Model}}} & \multicolumn{14}{c}{Language Consistency} \\
\cmidrule{2-15}          & AVG   & HRL   & URL   & EN    & DE    & FR    & ES    & RU    & ZH    & JA    & TH    & TE    & BN    & SW \\
    \midrule
    Qwen2.5-7B-Instruct & 68.4  & 92.6  & 24.3  & 100.0  & 80.0  & 94.8  & 98.0  & 95.2  & 100.0  & 87.6  & 38.8  & 16.0  & 2.4   & 40.0  \\
    xRFT  & 95.8  & 99.4  & 89.4  & 100.0  & 99.6  & 100.0  & 99.6  & 98.0  & 100.0  & 99.2  & 95.2  & 96.8  & 74.4  & \textbf{91.2} \\
    LIDR  & 52.3  & 69.1  & 15.2  & 100.0  & 27.6  & 72.8  & 98.0  & 80.4  & 98.8  & 37.2  & 11.2  & 1.1   & 0.8   & 47.6  \\
    MAPO  & 67.4  & 89.7  & 25.8  & 100.0  & 76.4  & 97.2  & 100.0  & 87.6  & 100.0  & 77.2  & 12.0  & 50.8  & 1.6   & 38.8  \\
    GRPO  & 17.8  & 14.4  & 2.3   & 100.0  & 1.6   & 3.2   & 1.6   & 0.4   & 74.4  & 5.2   & 1.2   & 0.4   & 1.2   & 6.4  \\
    polyGRPO & 9.6   & 0.2   & 1.1   & 99.6  & 0.0   & 0.4   & 0.0   & 0.0   & 0.4   & 0.4   & 0.4   & 2.4   & 0.0   & 1.6  \\
    polyGRPO$_{lang}$ w/ unconstrained prompt & 52.3 	& 58.4 	& 31.1  & 100  & 57.6  & 66.8  & 56.4  & 28.4  & 98.8  & 42.4  & 99.6  & 0.4  & 1.6  & 22.8  \\
    polyGRPO$_{lang}$ w/ language-constrained prompt & \textbf{99.1} & \textbf{100.0} & \textbf{97.4} & \textbf{100.0} & \textbf{100.0} & \textbf{100.0} & \textbf{100.0} & \textbf{100.0} & \textbf{100.0} & \textbf{100.0} & \textbf{100.0} & \textbf{100.0} & \textbf{100.0} & 89.6  \\
    \bottomrule
    \end{tabular}}%
  \label{apx_lang}%
\end{table}%

{We present only a preliminary investigation into the language consistency reward, which requires careful design. In particular, both the magnitude and the granularity (e.g., token- vs. sequence-level) of the reward may significantly influence the model’s attention to linguistic alignment. For instance, \citet{r1} define the reward as the proportion of tokens conforming to the target language at the token level. Magistral~\citep{Magistral} achieves notable gains in language consistency—and with minimal performance degradation—by employing a small amount of multilingual data together with a language consistency reward.}

\section{{Extending Polyglot Thinking to GSPO}}
Since our method primarily modifies the rollout procedure of GRPO—i.e., sampling reasoning traces in multiple languages—it can be readily adapted to GRPO variants such as DAPO~\citep{dapo} and GSPO~\citep{gspo}. Notably, GSPO elevates the optimization unit in reinforcement learning from the token level to the entire sequence level. Specifically, it replaces the per-token importance ratio  
$\frac{\pi_{\theta}^{i,t}}{\pi_{\theta_{\text{ref}}}^{i,t}}$  
with a sequence-level ratio, normalized by sequence length:

\begin{equation}
    \begin{aligned}
        s_i(\theta) =\left( \frac{\pi_{\theta}^{i}(o_i \mid p_{i}, q)}{\pi_{\theta_{\text{ref}}}^{i}(o_i \mid p_{i}, q)} \right)^{\frac{1}{|o_i|}} = \exp\left( \frac{1}{|o_i|} \sum_{t=1}^{|o_i|} \log \frac{\pi^{i,t}(o_{i,t} \mid p_{i}, q, o_{i,<t})}{\pi_{\theta_{\text{ref}}}^{i,t}(o_{i,t} \mid p_{i}, q, o_{i,<t})} \right)
    \end{aligned}
\end{equation}

Additionally, GSPO discards the KL-divergence penalty term used in GRPO. Consequently, the polyGSPO objective simplifies to:
\begin{equation}
\begin{aligned}
\mathcal{L}_{\text{polyGSPO}}(\theta) = \, &\mathbb{E}_{(q,a) \sim \mathcal{D}, \{o_i\}_{i=1}^n \sim {\pi_{\theta_{\text{ref}}}(o_i|p_i,q)}_{i=1}^n} \\ &\Bigg [
\frac{1}{n}\sum_{i=1}^n
\Bigg\{
\min\Bigg[s_i(\theta)\hat{A}_i, \text{clip}\Bigg(s_i(\theta), 1 - \epsilon, 1 + \epsilon\Bigg)\hat{A}_i\Bigg]
\Bigg\}\Bigg ]
\end{aligned}
\end{equation}

We implement both GSPO and polyGSPO based on Qwen2.5-7B-Instruct, using identical data, hyperparameters, and evaluation protocols as in prior experiments. Results on the MGSM benchmark (Table~\ref{polyGSPO}) show that:  GSPO underperforms GRPO-based methods—likely due to its sequence-level credit assignment being suboptimal for multi-step reasoning. Nevertheless, polyGSPO outperforms GSPO by +6.6\% and GRPO by +1.3\% in average accuracy, even without extensive hyperparameter tuning, confirming that \textbf{Polyglot Thinking consistently enhances reasoning capability—even under different RL optimization granularities.}


\begin{table}[htbp]
  \centering
  \caption{The results in MGSM benchmark based on GSPO and polyGSPO.}
    \resizebox{1\textwidth}{!}{\begin{tabular}{lccc|c|cccccc|cccc}
    \toprule
    \textbf{MGSM} & AVG   & HRL   & URL   & EN    & DE    & FR    & ES    & RU    & ZH    & JA    & TH    & TE    & BN    & SW \\
    \midrule
    \rowcolor[rgb]{ .906,  .902,  .902} \textit{Qwen2.5-7B-Instruct} &       &       & \multicolumn{1}{r}{} & \multicolumn{1}{r}{} &       &       &       &       &       & \multicolumn{1}{r}{} &       &       &       &  \\
    Base  & 67.2  & 75.7  & 48.8  & 89.2  & 73.6  & 74.8  & 79.2  & 78.8  & 80.0  & 68.0  & 73.6  & 36.4  & 68.0  & 17.2  \\
    xRFT  & 68.5  & 81.1  & 43.2  & \textbf{94.0}  & 81.6  & 78.0  & 84.4  & 83.6  & 85.2  & 73.6  & 69.6  & 25.2  & 54.4  & 23.6  \\
    LIDR  & 69.6  & 79.1  & 50.3  & 90.0  & 79.6  & 76.8  & 82.8  & 82.4  & 82.4  & 70.4  & 76.4  & 39.2  & 69.6  & 16.0  \\
    MAPO  & 66.3  & 75.8  & 47.4  & 84.8  & 78.4  & 76.4  & 79.2  & 78.8  & 74.8  & 67.2  & 74.4  & 33.2  & 62.8  & 19.2  \\
    GRPO  & 73.0  & 81.1  & 56.6  & 90.4  & 82.8  & 80.0  & 83.6  & 83.2  & 82.4  & 74.4  & 77.6  & 40.8  & 72.0  & 36.0  \\
    polyGRPO & \textbf{75.9} & \textbf{84.4} & \textbf{58.7} & \textbf{94.0}  & \textbf{86.0}  & \textbf{83.2}  & \textbf{88.8}  & \textbf{86.4}  & \textbf{84.8}  & \textbf{77.2}  & \textbf{81.2}  & \textbf{42.8}  & \textbf{74.8}  & \textbf{36.0}  \\
    \midrule
    GSPO	& 67.7 	& 76.8 	& 49.5 	& 85.6 	& 76.8 	& 73.2 	& 81.2 	& 78.8 	& 80.0 	& 70.8 	& 74.0 	& 30.0 	& 65.6 	& 28.4 \\
polyGSPO	& \textbf{74.3} 	& \textbf{83.9} 	& \textbf{55.5} 	& \textbf{92.4} 	& \textbf{84.4} 	& \textbf{84.0 }	& \textbf{84.8} 	& \textbf{88.4} 	& \textbf{85.2} 	& \textbf{76.4} 	& \textbf{76.4} 	& \textbf{42.0} 	& \textbf{69.6} 	& \textbf{34.0} \\
    \bottomrule
    \end{tabular}}%
  \label{polyGSPO}%
\end{table}%


\end{document}